\documentclass{article}

\usepackage[main, preprint]{neurips_2026}  

\usepackage[utf8]{inputenc}
\usepackage[T1]{fontenc}
\usepackage{hyperref}
\usepackage{url}
\usepackage{array}
\usepackage{booktabs}
\usepackage{amsfonts}
\usepackage{amsmath}
\usepackage{amssymb}
\usepackage{graphicx}
\usepackage{wrapfig}
\usepackage{nicefrac}
\usepackage{microtype}
\usepackage[table]{xcolor}

\usepackage{amsthm}
\newtheorem{definition}{Definition}
\newtheorem{proposition}{Proposition}

\usepackage{algpseudocode}
\usepackage{algorithm}

\algnewcommand\INPUT{\item[\textbf{Input:}]}%
\algnewcommand\OUTPUT{\item[\textbf{Output:}]}%
\algnewcommand\PARAM{\item[\textbf{Layer params:}]}%

\newcommand{\conciseparagraph}[1]{\noindent\textbf{#1.} }

\title{When Diffusion Breaks Constraints: Sequential Autoregressive Generation with RL and MCTS}

\author{%
Zirui Zhao\textsuperscript{1}\thanks{Corresponding to Zirui Zhao: \texttt{ziruiz@u.nus.edu}. Work was done at NUS.} \quad Boye Niu\textsuperscript{2} \quad Harold Soh\textsuperscript{3} \quad David Hsu\textsuperscript{3} \quad Wee Sun Lee\textsuperscript{3}\\
\textsuperscript{1}Salesforce AI Research \quad \textsuperscript{2}University of Sydney \quad \textsuperscript{3}National University of Singapore
}

\begin{document}
\maketitle

\begin{abstract}
Data-driven generative models excel in language and vision, but diffusion models often fail in constrained planning and design tasks, exhibiting severe constraint violations in engineering inverse design, molecular generation, multi-robot planning, and floorplan/scene synthesis even with projection or guidance. Such tasks combine hard-to-specify semantic goals with strict geometric or physical constraints (e.g., non-overlap, connectivity), yielding feasible solutions that lie on low-dimensional, small, and sometimes disconnected regions of the output space. This paper studies the failure mode through tangram generation from language, where seven fixed shapes must form a text-described silhouette while remaining connected and non-overlapping, and a simplified rectangle composition task with a learned bounding-box constraint. We find diffusion models struggle to satisfy constraints, consistent with difficulty generating samples near low-dimensional submanifolds. Motivated by locally feasible reparameterizations, we reformulate constrained generation as discrete autoregressive sequential generation. Reinforcement learning improves feasibility and task success, and Monte Carlo tree search quantifies the value of look-ahead when feasible regions shrink. Overall, the empirical, theoretical, and prior-work evidence points to a structural limitation of continuous density matching on this class of constrained-generation problems, and suggests sequential constraint-aware generation as a promising alternative.

\end{abstract}

\section{Introduction}

Data-driven generative models excel at producing realistic language, images, and videos, and diffusion models are increasingly the default choice when extending these successes to plan and design generation. However, striking failures arise when hard constraints are involved. In the \textsc{EngiBench} inverse-design benchmark, a conditional diffusion method violated problem constraints on nearly 100\% of the samples on the Beams2D and HeatConduction2D problems~\citep{felten2025engibench}. In molecular generation, the EDM diffusion model produces only $\approx 6\%$ stable molecules on GEOM-Drugs without bond-order post-processing~\citep{vignac2023midi}. In multi-robot motion planning, a diffusion planner drops from $78\%$ success rate on three robots planning in empty maps to $5\%$ success rate once obstacles are introduced, and to $0\%$ in the presence of six or more robots~\citep{liang2025projected}. Floorplan and scene synthesis diffusion models require bolted-on discrete branches, IoU penalties, or explicit repair modules to recover parallelism, non-overlap constraints, and connectivity~\citep{shabani2023housediffusion,tang2024diffuscene}. Post-hoc projection and classifier guidance have been used to mitigate the issues but often do not resolve the failures~\citep{christopher2024constrained,zampini2025training}.

Like language, images, and videos, these generation problems usually have properties such as style, class, and semantics that are difficult to specify and need to be learned from data. However, they often also have known constraints such as non-overlapping regions, connectivity, rigid placement, region containment, and physical feasibility, where a single violation would invalidate the output. As a result of these constraints, feasible solutions often lie on low dimensional submanifolds and/or in small and possibly disconnected parts of the output space. 

\begin{wrapfigure}{r}{0.5\columnwidth}
    \centering
    \vspace{-\intextsep}
    \includegraphics[width=\linewidth]{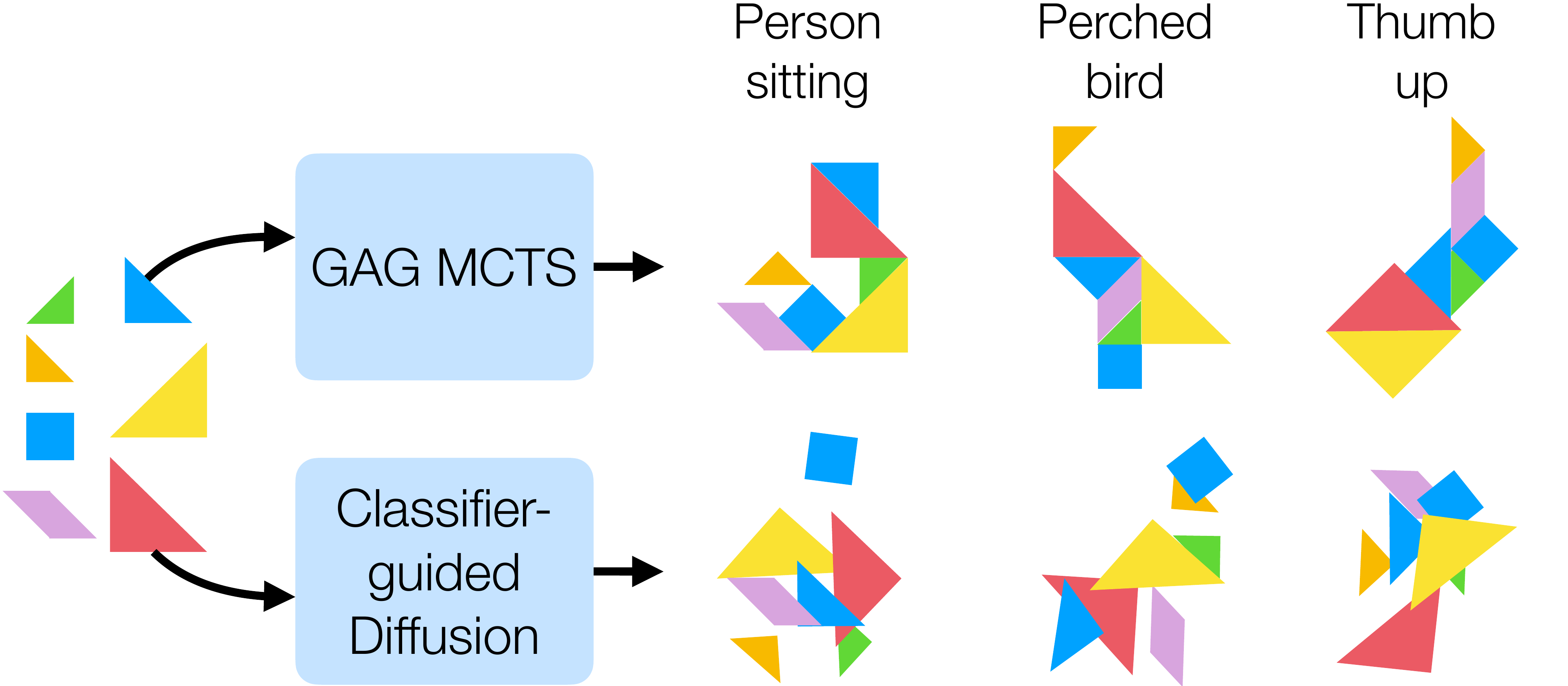}
    \vspace{-\intextsep}
    \caption{The Tangram generation task. The seven pieces are placed to form the target shape, described by the text such as ``perched bird''. Results from diffusion model and autoregressive model augmented with Monte Carlo tree search shown.}
    \label{fig:demo-front-page}
    \vspace{-0.5\intextsep}
    \end{wrapfigure}

We study these issues through a case study with easily visualized constraints and an abstract specification learned from data: \textbf{tangram generation} from language description. In tangram generation, seven known two-dimensional shapes must be composed into a silhouette matching a given free-form text cue, under the constraint that the shapes must be connected but cannot overlap, hence requiring both semantic understanding of language and geometric competence (see Figure~\ref{fig:demo-front-page}). To separate different aspects of the problem, we also study a relaxed problem, \textbf{rectangle composition}, where only rectangular shapes are used with non-overlap constraints and without requiring the shapes to be connected. To control problem difficulty, we provide a text specification of a bounding box that the rectangles must lie in as conditioning - this text-conditioned constraint must be learned from data, but is easy for humans to understand and visualize. 

We find that diffusion models have difficulty generating outputs that satisfy the tangram constraints, even with specifically designed constraint-energy guidance. To explain this, we draw on work of \citet{soh2026hallucination} showing that diffusion models have difficulty generating outputs close to specified low-dimensional submanifolds of the output space. We argue and show experimentally that feasible tangram solutions lie on such low-dimensional submanifolds. A natural response is to reparameterize the problem to remove redundant dimensions. A global reparameterization is difficult for tangram, but local reparameterization is tractable: given a subset of already placed pieces, the next piece can be generated by attaching it to an existing side. This motivates sequentially generating tangram pieces so that each new piece satisfies the constraints imposed by the current partial assembly. We further discretize the allowable position of the next generated piece, obtaining a standard discrete autoregressive generation problem.

Diffusion model is trained to generate all the variables simultaneously, exploiting the correlations among them. Autoregressive models, on the other hand, are conditioned on previously generated variables. To take future generation into account, reinforcement learning is often used with autoregressive models. In our study, we find that reinforcement learning is indeed effective in improving the performance of autoregressive constrained generation for the tangram and rectangle composition problems. Autoregressive generation methods are also easily used with tree search methods. We develop a Monte Carlo tree search method for the problem and find that it provides further improvements on top of the improvement provided by reinforcement learning. This is particularly the case for the rectangle composition problem when the bounding box is made smaller. When the bounding box is made smaller, the size of the feasible solution region becomes smaller and possibly disconnected. We extend the theoretical arguments in \citet{soh2026hallucination} showing diffusion models have difficulties in such problems, and show that the use of search, e.g. through Monte Carlo tree search, provides a fairly effective way for autoregressive generation to use additional computation to improve performance.

Through this case study, we aim to clarify why continuous generators struggle on hard constrained-generation problems and why local, sequential reparameterisations can help. While we only study tangram generation and rectangle composition, the same geometric issue appears in several practical constrained-generation domains, suggesting that the lesson is broader than these two testbeds.

\section{Related Works}

\conciseparagraph{Empirical failure of generative models on constrained tasks}\label{sec:related_empirical}
Continuous-support generative models frequently violate hard constraints across domains. Diffusion baselines violate constraints on nearly all samples in the \textsc{EngiBench} inverse-design benchmark~\citep{felten2025engibench} and produce only $\approx 6\%$ stable molecules on GEOM-Drugs~\citep{vignac2023midi}. Multi-robot diffusion planners degrade sharply with obstacles and robot count~\citep{liang2025projected}, and floorplan/scene synthesis models~\citep{shabani2023housediffusion,tang2024diffuscene} require bolted-on discrete branches or repair modules to recover non-overlap and connectivity. These failures share a structural pattern: feasible outputs lie on low-dimensional or disconnected regions of a much larger ambient space.

\conciseparagraph{Constraint-aware generative pipelines}
A parallel line of work tries to make generative models satisfy hard rules by \emph{augmenting} them at training or inference time. Geometric-loss GANs~\citep{zeng2021enforcing}, factor-graph layouts~\citep{dupty2024constrained}, primitive-plus-constraint CAD models such as SketchGen~\citep{para2021sketchgen}, dual-discriminator GANs~\citep{almasri2020shape}, and parameter-efficient mechanical-design models~\citep{padula2024generative} incorporate constraints into the generative pipeline to varying degrees. For diffusion specifically, post-hoc projection (mirror mapping~\citep{liu2023mirror}, optimisation-based sampling~\citep{christopher2024constrained}) and training-free heuristics~\citep{zampini2025training} recover partial feasibility but remain approximate and frequently break under novel configurations. Recent MCTS-guided diffusion planning~\citep{yoon2025monte} uses rewards to steer diffusion trajectories, but the underlying action space is still continuous and hard constraints remain implicit. These methods are valuable, but as \citet{soh2026hallucination} make precise (Section~\ref{sec:theory}), the underlying failure is a geometric property of continuous density matching on small and often measure-zero sets, so symptomatic fixes cannot close the gap in general.

\conciseparagraph{Sequential discrete generation and search for combinatorial problems}
When feasibility is a combinatorial predicate, discrete autoregressive generation with reinforcement learning and tree search is a natural alternative. AlphaZero~\citep{silver2018alphazero}, MuZero~\citep{schrittwieser2020mastering}, and Gumbel MuZero~\citep{danihelka2022policy} show that policy-value search can navigate large rule-constrained action spaces, while PPO~\citep{schulman2017proximal} provides a standard RL post-training recipe for autoregressive policies. However, discretisation alone does not make the problem easy. A full-state discrete diffusion model over seven-piece tangram layouts would denoise in an exponentially large joint space, where most states violate constraints. Our method instead uses the discrete structure locally: it factorises generation into placement actions conditioned on the current partial assembly, and uses search to reason about future completions. We validate this sequential-search recipe on tangram assembly~\citep{clark1986referring,ji2022abstract} and rectangle packing, two settings where feasibility is measurable and the combinatorial structure is explicit.

\section{Problem Formulation}
\label{sec:problem_formulation}

We formalise the class of constrained-generation problems introduced in Section~1 using planning terminology. Let $S$ be the ambient state space of partial or complete compositions, and let $A(s)$ be the set of admissible composition operations available at state $s$. A terminal state is valid only if it satisfies hard constraints $C(s)$, such as geometric constraints in tangram assembly, syntactic constraints in text generation, or logical constraints in reasoning tasks. Given an initial state $s_0$, a transition function $T(s,a)$, and a specification $y \in Y$, the constrained composition problem is to find an action sequence whose terminal state is both feasible and goal-satisfying: \begin{wrapfigure}{r}{0.4\columnwidth}
    \centering
    \vspace{-0.5\intextsep}
    \includegraphics[width=0.4\columnwidth]{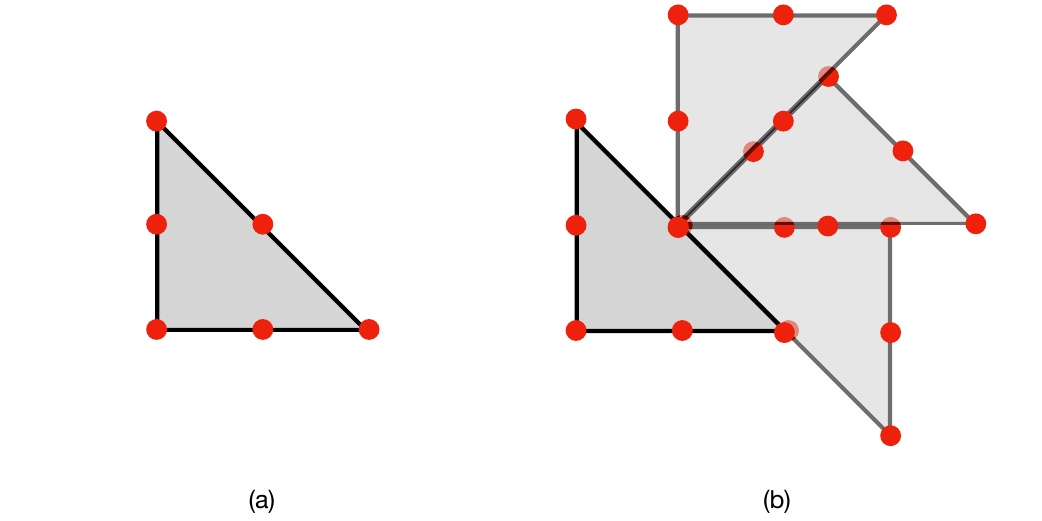}
    \vspace{-1.5\intextsep}
    \caption{Actions in Tangram assembly. Pieces can be placed when anchor points align with rotations without overlapping.}
    \label{fig:tangram_demo}
    \vspace{-\intextsep}
\end{wrapfigure}\begin{align}
    \text{find}~& a_{1:T}, \quad a_t \in A(s_{t-1}), \quad s_t = T(s_{t-1}, a_t)\notag\\
    \text{s.t.}~& C(s_T) = \text{True}, \quad G(s_T, y) = \text{True}.
\end{align}
$C(s_T)$ denotes the domain-specific constraints that must be satisfied at completion, and the goal function $G(s_T, y)$ determines whether the terminal state fulfils the requirements specified by $y$. Intermediate states may be partial and need not themselves satisfy the terminal constraints.

This formulation encompasses various constrained composition tasks. In tangram assembly, seven fixed pieces must be arranged to match a text-specified concept under geometric constraints: $S$ is the space of partial or complete layouts using two small triangles, one medium triangle, two large triangles, one square, and one parallelogram; $Y$ is the space of natural-language descriptions of shapes; and the terminal constraints $C(s_T)$ require \emph{(1) the contact graph over all seven pieces is connected, (2) no overlaps, (3) all seven pieces are used}. The connectivity constraints are active equalities that confine feasible configurations to a stratified subset of the ambient configuration space, where each stratum (indexed by a combinatorial contact pattern) is a smooth submanifold (Appendix~\ref{app:theory_apply}). For each assembly step, one unplaced piece is placed adjacent to an already placed piece at an anchor-aligned pose (Fig.~\ref{fig:tangram_demo}), with rotations discretised in $45^\circ$ increments and piece-specific flip states. In a rectangle composition, the action places one rectangle on an integer grid with $0^\circ$ or $90^\circ$ orientation inside a target region. The goal function $G(s_T, y)$ evaluates whether the final state matches the goal specification $y$.

\section{Feasible-Set Barriers for Continuous Generators}
\label{sec:theory}

We now formalise why continuous generators struggle on constrained-generation tasks, building on ideas from~\citet{soh2026hallucination}. The key objects are the \emph{feasible set} determined by the hard constraints and its \emph{relative volume} inside the ambient layout space.

Let $\mathcal{X}_y$ be the continuous layout space for task instance $y$ and let $\mathrm{viol}_y(x)\geq 0$ be a \emph{geometric} violation score (overlap and connectivity gap; semantic mismatch is handled separately by the reward and can only shrink the feasible set further, making the bounds below conservative).

\begin{definition}[Tolerant feasible set]
For tolerance $\epsilon\geq 0$, define
$\mathcal{F}_y^\epsilon:=\{x\in\mathcal{X}_y:\mathrm{viol}_y(x)\leq\epsilon\}$.
The exact feasible set is $\mathcal{F}_y:=\mathcal{F}_y^0$.
\end{definition}

\begin{definition}[Relative feasible mass]
Assuming $\mathcal{X}_y$ is bounded with uniform reference density $p_y$, define
$\mu_y(\epsilon):=\mathrm{Vol}(\mathcal{F}_y^\epsilon)/\mathrm{Vol}(\mathcal{X}_y)$.
\end{definition}

Let a continuous generator induce a density $p_{\theta,\beta}(x\mid y)$, where $\theta$ are model parameters and $\beta$ an optional guidance scale. Define the generator's \emph{concentration factor} on the feasible set:
\begin{equation}
K_{\theta,\beta}(y,\epsilon) \;:=\; \operatorname{ess\,sup}_{x\in\mathcal{F}_y^\epsilon}\frac{p_{\theta,\beta}(x\mid y)}{p_y(x)}.
\end{equation}

\begin{proposition}[Feasible-mass bound]
\label{prop:feasible_mass}
For any one-shot continuous generator with density $p_{\theta,\beta}(x\mid y)$,
\begin{equation}
S_{\theta,\beta}(y,\epsilon) \;:=\; \Pr\nolimits_{x\sim p_{\theta,\beta}}[x\in\mathcal{F}_y^\epsilon]
\;\leq\; K_{\theta,\beta}(y,\epsilon)\;\mu_y(\epsilon).
\label{eq:feasible_mass_bound}
\end{equation}
\end{proposition}
The proof follows from multiplying and dividing by $p_y$ and pulling out the essential supremum; see Appendix~\ref{app:theory_statements}.
If $\mu_y(\epsilon)$ is small, a generator must supply a correspondingly large concentration factor to achieve constant validity. Guidance can increase $K_{\theta,\beta}$ by steering samples toward low-violation regions, but it cannot change $\mu_y$, which is the core feasible-set barrier. In Section~\ref{sec:analysis} we show that $\mu_y$ is indeed small for both of our tasks through \emph{contact precision} in tangram and \emph{pose-volume depletion} in rectangle composition, and that the qualitative regimes are consistent with the empirical performance hierarchy.

The feasible-mass bound suggests a natural remedy: instead of matching a density in the full ambient space, parameterise the feasible set locally, one piece at a time. Given a committed prefix, the feasible placements of the next piece can be enumerated directly. Discretising this set reduces generation to a standard discrete autoregressive problem, for which strong RL and tree-search algorithms exist (PPO~\citep{schulman2017proximal}; AlphaZero-style policy improvement and its Gumbel variant~\citep{silver2018alphazero,schrittwieser2020mastering,danihelka2022policy}). Discrete structure makes the tube-density requirement vacuous, and tree search directly addresses the sequential nature of the feasible-mass decomposition by choosing actions that preserve future feasibility.

\section{Generative Adversarial Gumbel MCTS}
\label{sec:method}

Reinforcement learning is commonly used for autoregressive generation, although obtaining an appropriate reward function is often difficult. We use PPO~\citep{schulman2017proximal} as one RL baseline. In this section, we focus on a generative version of Monte Carlo tree search, Gumbel MCTS, and on how we improve the learned reward function through adversarial training.

\begin{wrapfigure}{r}{0.55\columnwidth}
    \vspace{-\intextsep}
    \includegraphics[width=0.55\columnwidth]{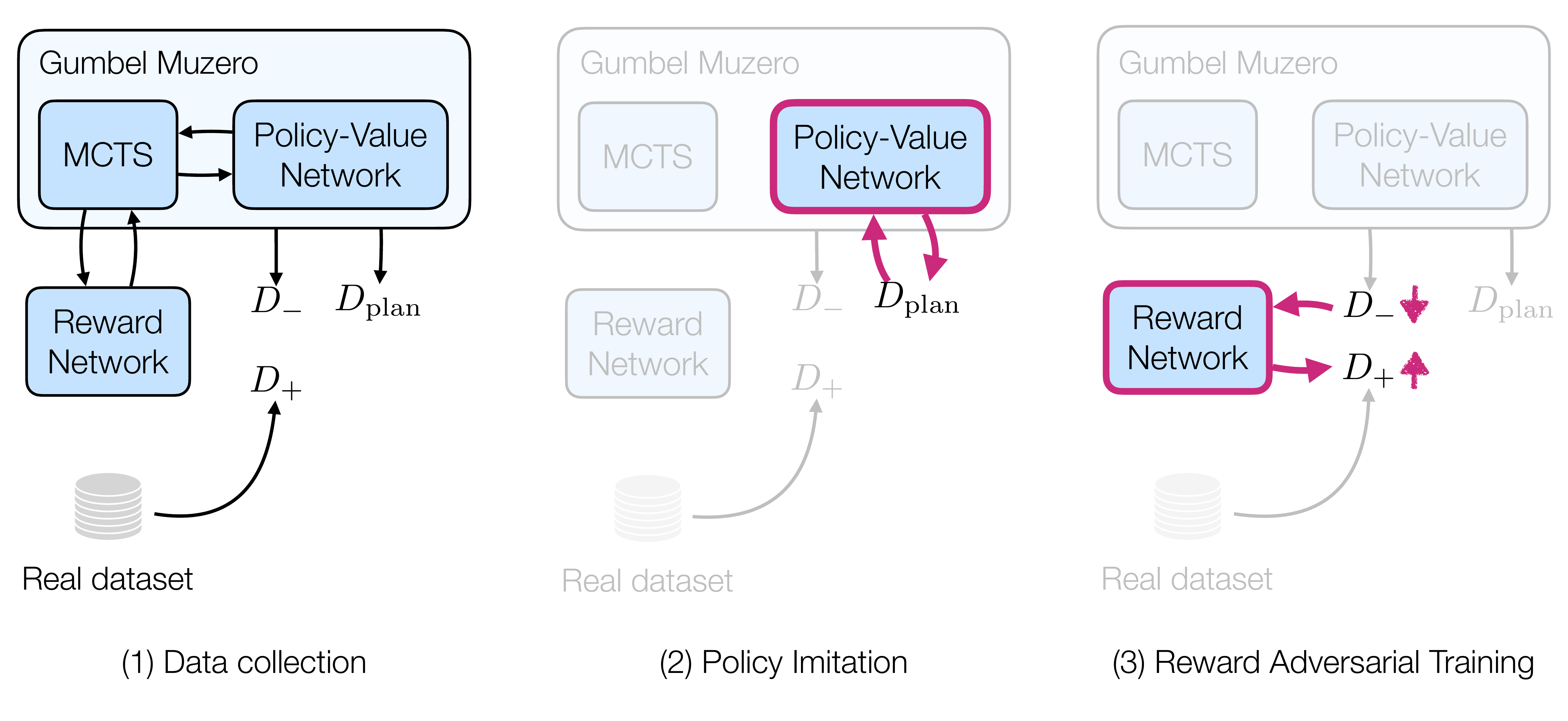}
    \vspace{-2\intextsep}
    \caption{Generative Adversarial Gumbel MCTS. Black arrows carry outputs between modules; purple arrows denote gradient updates.}
    \label{fig:ch5_overview}
    \vspace{-0.5\intextsep}
\end{wrapfigure}

The per-step action space is discretised as in Section~\ref{sec:problem_formulation}. A pairwise action mask filters anchor-alignment and immediate overlap violations between the newly placed piece and the already placed pieces (Appendix~\ref{app:action_mask}), so the policy mostly scores locally feasible continuations; the mask does not certify that the remaining pieces can still complete the assembly. We use the root sequential-halving search introduced in Gumbel MuZero~\citep{danihelka2022policy}, but transitions are advanced by the exact geometric simulator rather than by a learned dynamics model. We therefore refer to the resulting search operator as Gumbel MCTS (Appendix~\ref{app:gumbel_muzero}): its policy-value network supplies action logits and value estimates for MCTS, and we train both by imitating MCTS-produced trajectories (Fig.~\ref{fig:ch5_overview}).

Since the reward is far easier to learn than the composition itself~\citep{ji2022abstract}, we use a fine-tuned CLIP reward as the discriminator and train it adversarially: MCTS rollouts under the current policy produce negatives $D_-$, labelled configurations serve as positives $D_+$, and we alternate policy improvement on the planning trajectories $D_\mathrm{plan}$ with reward refinement on $D_+\cup D_-$ (details in Section~\ref{sec:adversarial} below).

\begin{wrapfigure}{r}{0.5\columnwidth}
    \centering
    \vspace{-\intextsep}
    \includegraphics[width=0.5\columnwidth]{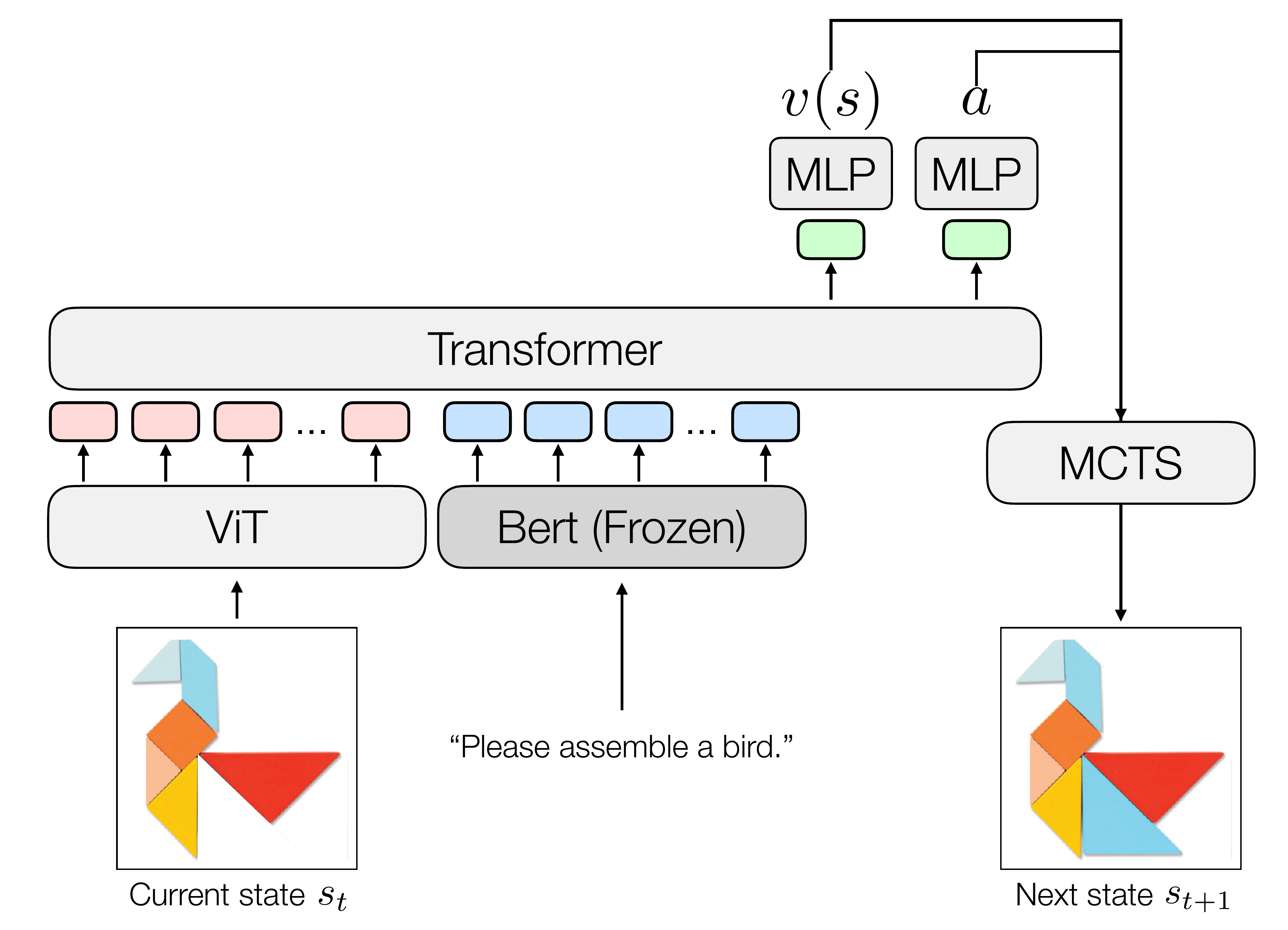}
    \vspace{-\intextsep}
    \caption{The policy value network. We use a pre-trained Vision Transformer (ViT) and Bert, together with a transformer that fuses the vision and language features, to form the main body of the network. We use a single network to predict the value and next action, as suggested by the AlphaZero paper. The last two tokens are decoded as a value and action logits after a linear layer. }
    \label{fig:policy_value_network}
    \vspace{-3\intextsep}
\end{wrapfigure}

\conciseparagraph{Policy-value network}
Following vision-language-action architectures in robotics, we fuse a pre-trained ViT image encoder and a fixed BERT text encoder through a small transformer; two output tokens decode (via a linear head) to action logits and a scalar value (Fig.~\ref{fig:policy_value_network}). We keep BERT fixed and only fine-tune ViT and the fusion transformer, since KiloGram is small and a fixed text encoder stabilises the text features under limited data. The network is trained to imitate MCTS-produced trajectories and predict their return, and is used inside MCTS as the prior and rollout value. Further design rationale is in Appendix~\ref{app:ch5_implementation_details}.

\conciseparagraph{Reward network}
We use the pre-trained vision-language model CLIP as the reward network. The reward network takes the tangram configuration and text description as inputs and outputs a scalar score for their semantic match.

\subsection{Gumbel MCTS with Adversarial Reward Refinement}\label{sec:adversarial}
First, we use the training data in the Tangram dataset to fine-tune the pre-trained CLIP model using contrastive learning. It can achieve near-human accuracy on the validation and test splits, as reported in the original KiloGram paper. However, during Gumbel MCTS training, we found that generated tangram configurations can receive high reward even when the silhouette does not actually match the text description. Thus, the fine-tuned reward model is not robust to search-generated near misses and can be fooled by false-positive instances.

To address this, we train the reward adversarially against MCTS-generated negatives. Let $r_\phi$ denote the CLIP reward with parameters $\phi$, $s_i$ a tangram image and $\ell_i$ its text description. We treat labelled pairs $(\ell_i, s_i^+)\in D_+$ as positives and MCTS-generated pairs $(\ell_i, s_i^-)\in D_-$ as negatives and minimise the binary cross-entropy
\begin{align}
    \mathcal{L}_\mathrm{pref} \;=\; -\frac{1}{N}\!\!\sum_{(\ell_i, s_i^+)\in D_+}\!\!\!\! \log \sigma(r_\phi(\ell_i, s_i^+))
    \;-\; \frac{1}{N}\!\!\sum_{(\ell_i, s_i^-)\in D_-}\!\!\!\! \log\big(1 - \sigma(r_\phi(\ell_i, s_i^-))\big),
\end{align}
where $\sigma$ is the sigmoid and $N$ the batch size.
To avoid catastrophic forgetting, we combine $\mathcal{L}_\mathrm{pref}$ with the standard CLIP contrastive loss $\mathcal{L}_\mathrm{cont}$ on $D_+$: within each batch, every image is pushed to match only its paired caption (and vice versa), so the reward continues to reinforce the original image--text correspondences while the preference term sharpens its discrimination against MCTS-generated negatives. The total reward objective is $\mathcal{L}_\mathrm{pref} + \lambda \mathcal{L}_\mathrm{cont}$, with $\lambda$ a hyperparameter; the full InfoNCE~\citep{oord2018representation} formula is given in Appendix~\ref{app:ch5_implementation_details}.

We train the policy and the reward jointly in a generative-adversarial loop shown in Fig.~\ref{fig:ch5_overview}. It alternates three steps: (i) \emph{Plan}: starting from initial states sampled from the positive dataset $D_+$, the current policy-value network $\pi_\theta$ is rolled out under Gumbel MCTS; the search produces planning trajectories $D_\mathrm{plan}$ and completed configurations $D_-$. (ii) \emph{Improve the policy}: $\pi_\theta$ is fine-tuned on $D_\mathrm{plan}$, using either PPO or MCTS-imitation targets. (iii) \emph{Refine the reward}: $r_\phi$ is fine-tuned on $D_+$ as positives and $D_-$ as negatives, with the combined loss $\mathcal{L}_\mathrm{pref} + \lambda \mathcal{L}_\mathrm{cont}$. The binary term pushes the reward to discriminate near-miss generations, while the contrastive term guards against catastrophic forgetting. The two halves are coupled: the policy searches against a reward that becomes stricter every iteration, and the reward sees an ever-harder stream of MCTS negatives, yielding mutual improvement until the generated distribution closely matches $D_+$. Adversarial training is sensitive to hyperparameters, so we use a small gradient step for $r_\phi$ and a relatively large MCTS simulation budget per iteration.

\section{Experiments and Analysis}
\label{sec:experiments}

We evaluate the sequential on-manifold Monte Carlo tree search recipe (instantiated as GAG MCTS, Section~\ref{sec:method}) against diffusion and autoregressive baselines on two concrete problems from the class of Section~\ref{sec:problem_formulation}. \textbf{Tangram assembly} (Section~\ref{sec:exp_tangram}) places hard non-overlap and connectivity constraints on a 21-dimensional configuration space and adds a weakly-specified text goal; \textbf{rectangle composition} (Section~\ref{sec:exp_rect}) strips the semantic component and varies difficulty via the piece-to-region fill ratio~$r$. Section~\ref{sec:analysis} returns to the theory and shows that the observed pattern across baselines and regimes is consistent with the feasible-mass analysis of Section~\ref{sec:theory}.

\subsection{Tangram Assembly}
\label{sec:exp_tangram}

The task is to place the seven tangram pieces into a configuration whose silhouette matches a free-form text description. It simultaneously stresses geometric feasibility (non-overlap and connectivity on a 21-dimensional configuration space) and semantic alignment under weak supervision, exercising both sides of the constrained-generation class.

\subsubsection{Experiment setup}\label{ch5:exp_tangram_setup}

\begin{wrapfigure}{r}{0.55\columnwidth}
    \vspace{-5\intextsep}
    \centering
    \includegraphics[width=0.55\columnwidth]{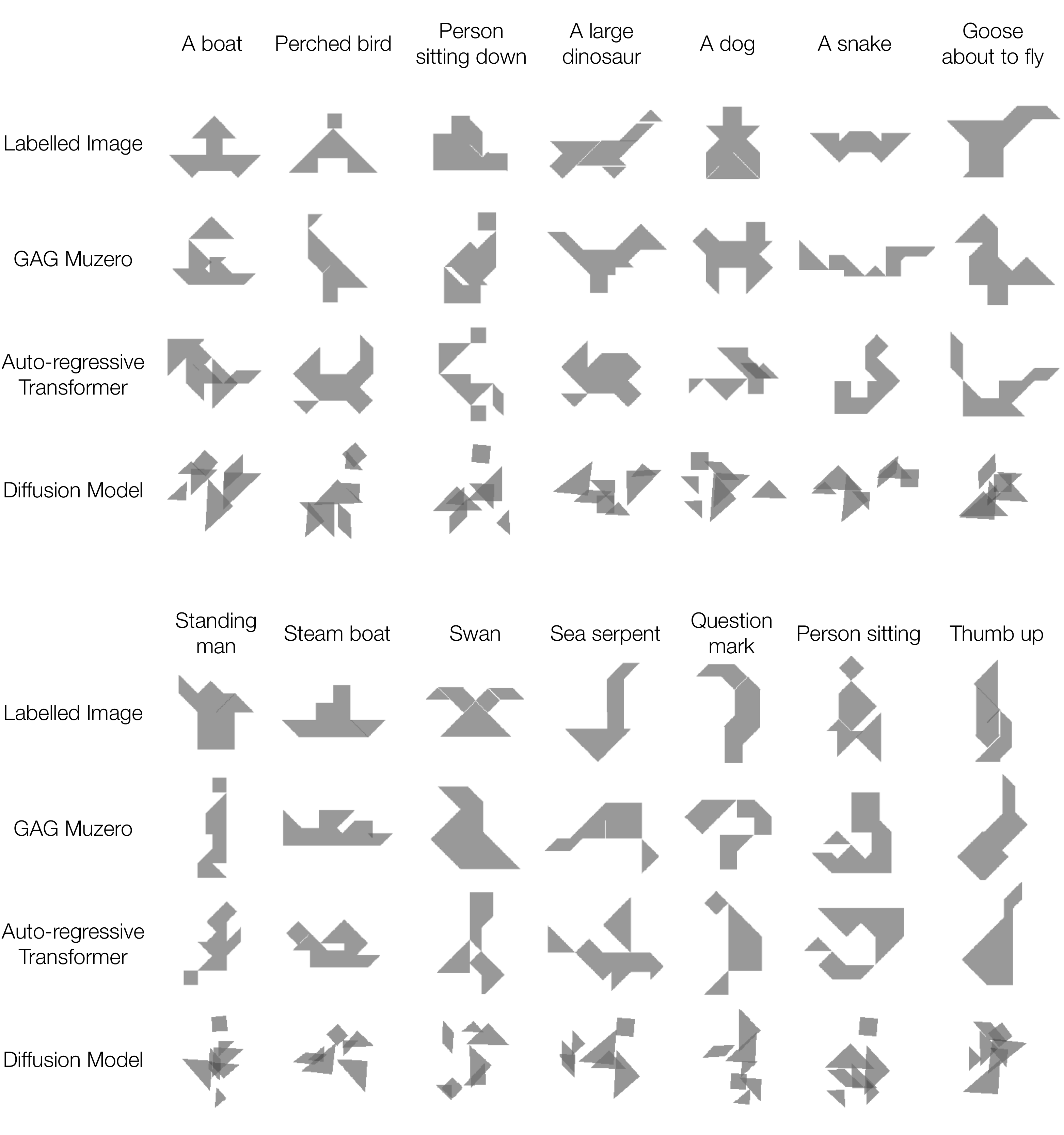}
    \vspace{-2\intextsep}
    \caption{Dataset and generated tangram configurations; GAG MCTS vs.\ main baselines.}
    \label{fig:tangram_demo_fig}
    \vspace{-\intextsep}
\end{wrapfigure}

\conciseparagraph{Dataset} We use the KiloGram dataset~\citep{ji2022abstract} for the experiment. The KiloGram dataset contains thousands of tangram pieces with tens of thousands of text annotations. The dataset is initially designed for abstract visual concept comprehension, i.e., tangram classification. In our study, we modify the dataset for constrained generation, which is empirically much harder than recognition/classification, even with discretisation in the action space. Some example images with annotations in the dataset are shown in Fig~\ref{fig:tangram_demo_fig}. 

\begin{table}[htbp]
\centering\small
\setlength{\tabcolsep}{4pt}
\renewcommand{\arraystretch}{1.15}
\caption{Methods studied in Tangram Assembly. ``FT CLIP'' denotes CLIP fine-tuned on KiloGram; ``Adv.'' denotes adversarial reward refinement.}
\label{tab:baselines}
{\rowcolors{2}{white}{gray!12}%
\begin{tabular}{@{}m{0.22\linewidth} m{0.19\linewidth} m{0.12\linewidth} m{0.19\linewidth} m{0.20\linewidth}@{}}
\toprule
Baseline & Action selection & Action mask & Reward model & Training \\
\midrule
Random & Uniform & None & -- & -- \\
Energy-guided diffusion ($\beta$)~\citep{yu2023freedom} & DDPM + energy-guidance gradient & None & -- & Supervised \\
Autoregressive & Transformer logits & None & -- & Supervised \\
Autoregressive + mask & Transformer logits & Pairwise & -- & Supervised \\
PPO + mask (+Adv) & Transformer logits & Pairwise & FT CLIP (+Adv.) & SFT $\to$ PPO \\
Gumbel MCTS w/o GA & MCTS & Pairwise & FT CLIP (no Adv.) & MCTS imitation \\
GAG-MCTS w/o pre-trained CLIP & MCTS & Pairwise & From scratch + Adv. & MCTS imitation + GA \\
GAG MCTS  & MCTS & Pairwise & FT CLIP + Adv. & MCTS imitation + GA \\
\bottomrule
\end{tabular}%
}
\end{table}

\conciseparagraph{Methods studied}
Table~\ref{tab:baselines} summarises the seven methods. All autoregressive, PPO, and search methods share the same \emph{pairwise} action mask: a precomputed anchor-aligned geometric filter that catches immediate overlaps but does not certify future feasibility (Appendix~\ref{app:action_mask}); search methods gain their advantage from look-ahead on top of this mask. Energy-guided diffusion~\citep{yu2023freedom} uses no discrete mask but encodes similar geometric preferences via a differentiable energy function with guidance scale $\beta$. A full-state discrete diffusion formulation would face an exponential joint state space: even a conservative count gives $7!\times100^6\approx5\times10^{15}$ ordered assemblies, mostly infeasible (Appendix~\ref{app:action_mask}). Reward-based baselines share the fine-tuned CLIP backbone to isolate the effect of pretraining and adversarial refinement.

\begin{wraptable}{r}{0.6\columnwidth}
\vspace{-0.5\intextsep}
\centering\small
\setlength\tabcolsep{2.6pt}
\caption{Results on Tangram Assembly. $t$ means the temperature of sampling, $\beta$ is the energy-guidance scale, and $g$ means the Gumbel scale in Gumbel MCTS. }\label{tab:ch5_main}
\vspace{0.5em}
{\rowcolors{2}{white}{gray!12}%
\begin{tabular}{@{}lccccc@{}}
\toprule
                                   & FID$\downarrow$ & FID$_\text{clip}$$\downarrow$ & Pre $\uparrow$ & Rec $\uparrow$ & Val(\%)$\uparrow$ \\ \midrule
Random                              & 35.8 & 0.71 &  0.380    & 0.231 & 0.3 \\
Diffusion ($\beta = 0$)                 & 34.9 & 0.54 &   0.350    &  0.203  & 0.0 \\
Diffusion ($\beta = 0.1$)                 & 34.4 & 0.54 &   0.347    &  0.211  & 0.0 \\
Diffusion ($\beta = 1.0$)                 & 34.0 & 0.53 &   0.368    &  0.239  & 1.6 \\
Diffusion ($\beta = 10.0$)                & 33.2 & 0.59 &   0.370    &  0.248  & 4.3 \\
Auto-reg ($t=1.0$)                & 28.6  & 0.59  &  0.580 & 0.410 & 54.2\\
Auto-reg ($t=0.5$)                 & 24.5 & 0.52  &   0.571 & 0.427  & 59.3\\
Auto-reg (Greedy)                  & 23.2 & 0.48 &   0.564    &  0.424   & 61.1\\
Auto-reg+mask ($t=1.0$)            & 25.0  & 0.57  &  0.553 & 0.409  & 87.9\\
Auto-reg+mask ($t=0.5$)            & 23.1 & 0.50  &   0.575 & 0.430  &  88.1\\
Auto-reg+mask (Greedy)             & 22.0 & 0.48 &   0.568    &  0.423  &  86.4\\
PPO+mask (Greedy)             & 19.7  & 0.45 &  0.553  &  0.311  &  99.3 \\
PPO+adv+mask (Greedy)             & 18.6 & 0.42 &    0.587   &  0.402  & 99.3 \\
Gumbel MCTS w/o GA  & 22.9 & 0.46  &  0.569    &  0.412 & 98.5 \\
\begin{tabular}[c]{@{}l@{}}GAG MCTS w/o\\pre-trained CLIP\end{tabular}       & 32.3 & 0.53  &  0.503    &  0.491 & 98.5\\
GAG MCTS ($g=1.0$)           & 18.2  & 0.41  &   0.576 &  0.459  & 99.3 \\ 
GAG MCTS ($g=0.5$)           & 17.6  &  0.40  &  0.585 &  0.451  & 98.5 \\ 
GAG MCTS (Greedy)           & 16.8  &  0.39  &  0.591 &  0.444  & 99.3 \\ \bottomrule
\end{tabular}%
}
\vspace{-\intextsep}
\end{wraptable}

\conciseparagraph{Metric}
Our \emph{primary} metric is the percentage of valid configurations (all seven pieces placed, no pairwise overlap, connected), since this is the hard-constraint property the task requires; FID (with both a pre-trained InceptionV3 and a KiloGram-fine-tuned CLIP) and the precision/recall metric of~\citet{kynkaanniemi2019improved} are \emph{supportive} indicators of likelihood, fidelity, and diversity. Full metric definitions are in Appendix~\ref{app:metrics}.

\conciseparagraph{Implementation details}
We use JAX to create and vectorise the Tangram assembly task. The MCTS is adapted from the Mctx package, which uses JAX to conduct parallelised batch training. The search experiments measure test-time-compute scaling on top of the same discrete generator family; runtime and hardware details are reported in Appendix~\ref{app:inference_cost}. For more details, please refer to Appendix~\ref{app:ch5_implementation_details}.

\subsubsection{Main Result}

Table~\ref{tab:ch5_main} reports validity (percentage of outputs satisfying all hard constraints), FID for distributional fidelity, and precision/recall for mode coverage. We highlight the main findings below.

\textbf{Search outperforms continuous-output baselines.}
Energy-guided diffusion attains only $0$--$4\%$ validity even with strong guidance scale; the autoregressive transformer reaches $54$--$88\%$ depending on masking; PPO with the same mask lifts validity to $99.3\%$ but leaves FID and precision/recall clearly worse than search (FID $18.6$ vs.\ $16.8$ for GAG MCTS greedy, recall $0.40$ vs.\ $0.44$). GAG MCTS reaches near-saturation validity and has the best FID/precision. Section~\ref{sec:analysis} provides a qualitative analysis via the feasible-mass bound.

\textbf{Reward-model ablations.} Removing adversarial refinement (\emph{Gumbel MCTS w/o GA}) degrades FID and precision/recall and leaves the reward easily exploitable by wrong configurations. Removing the CLIP pretraining (\emph{GAG MCTS w/o pre-trained CLIP}) is worse: the reward CLIP then recognises only $\approx 30\%$ of KiloGram validation shapes and the resulting generator has the lowest precision among search-based methods, confirming that both pretraining and adversarial refinement are important under data scarcity.

\textbf{Gumbel noise trades precision for recall.} Sweeping the Gumbel scale from greedy to $g\!=\!1.0$ slightly lowers precision ($0.591\to 0.576$) but raises recall ($0.444\to 0.459$), providing a knob for diversity. Recall nonetheless remains below precision, indicating residual mode collapse, which is typical of GAN-style training with limited data (Fig.~\ref{fig:tangram_demo_fig}).

\conciseparagraph{Human study}
As a qualitative check on whether FID and precision/recall track what humans perceive as a good match, we run a small forced-choice study. Five participants each view 50 image pairs, where each pair shows a ground-truth KiloGram configuration alongside a generation (GAG MCTS ($g\!=\!0.5$) or the autoregressive baseline with $t\!=\!0.5$), randomly ordered left/right, with the prompt ``Which image better matches the description below?'' Participants preferred GAG MCTS over ground truth on $39.2\!\pm\!4.4\%$ of pairs, vs.\ $4.4\!\pm\!2.6\%$ for the autoregressive baseline, where the error bars are standard deviations across participants. A paired $t$-test confirms the difference is statistically significant ($t(4)=11.33$, $p<0.001$, Cohen's $d=5.07$). This small study provides a supportive semantic-quality check for the distributional metrics. Details are in Appendix~\ref{app:human_study}.

\subsection{Rectangle Composition}
\label{sec:exp_rect}

Tangram mixes hard constraints with goal ambiguity and does not allow us to vary the difficulty of the problem. To study these issues, we use a relaxed variant of the problem: arrange a fixed set of rectangles inside a text-specified rectangular region under a non-overlap constraint, with the connectivity constraint removed. The target region serves as a text-derived geometric constraint, and difficulty is controlled by the number/size of rectangles relative to the container dimensions. This lets us test whether issues in the tangram problem persist once goal ambiguity difficulty is reduced, the connectivity constraint is removed, and how it scales with problem difficulty.

\begin{wraptable}{r}{0.55\columnwidth}
\vspace{-1.5\intextsep}
\centering\small
\setlength\tabcolsep{2.6pt}
\caption{Results on Rectangle Packing (success rate \% in the first three columns, valid rate \% in the last column). $t$: the temperature of sampling; $\beta$: the energy-guidance scale; $g$: the Gumbel scale in Gumbel MCTS. }\label{tab:ch5_rectangle}
\begin{tabular}{@{}lcccc@{}}
\toprule
                                    & Easy & Hard & Average & Valid \\ \midrule
Diffusion ($\beta = 0$)                 &  88.50  & 21.00 & 54.75 & 69.75 \\
Diffusion ($\beta = 0.1$)                 &  89.00  & 21.50 & 55.25 & 70.50 \\
Diffusion ($\beta = 1.0$)                 &  93.50 & 19.50 & 56.50 & 75.50 \\
Diffusion ($\beta = 10.0$)                 &  97.00 & 15.50 & 56.25 & 82.75 \\
Auto-reg+Mask ($t=1.0$)                 &  55.50  & 2.00 & 28.75 & 84.25 \\
Auto-reg+Mask ($t=0.5$)                 & 59.00 & 1.00 & 30.00 & 85.50 \\
Auto-reg+Mask (Greedy)                  &  51.50 &  1.50  & 26.50 & 88.50 \\
PPO+Mask (Greedy)                  &89.50 & 13.00  &   51.25 & 94.50 \\
PPO+Adv+Mask (Greedy)            & 94.00  &  30.00  & 62.00  & 95.50 \\
GAG MCTS ($g=1.0$)          &  95.50  &  50.50 & 73.00  & 98.50 \\ 
GAG MCTS ($g=0.5$)          &  95.00 & 53.50  &  74.25   & 97.75 \\ 
GAG MCTS (Greedy)          &  96.00 &  51.00 &   73.50  & 95.25 \\ \bottomrule
\end{tabular}
\vspace{-0.5\intextsep}
\end{wraptable}


\conciseparagraph{Dataset \& Metric}
We generate synthetic rectangle-packing problems from three piece types ($1\!\times\!2$, $2\!\times\!3$, $1\!\times\!5$) with axis-aligned rotations. For each problem, we sample a target region ($3\!\times\!3$ to $12\!\times\!12$), find a complete tiling by randomised backtracking, and keep a random $10$--$100\%$ subset of the pieces to produce packing instances of varying difficulty (every instance has at least one valid solution). The dataset contains $2{,}000$ training, $100$ validation, and $400$ test instances. Training and test draw pieces from different inventory mixes by design, so a solver must generalise beyond the exact training inventory rather than memorise it; the exact counts are in Appendix~\ref{app:ch5_rectangle_gen}. Difficulty is summarised by the fill ratio $r$=(total piece area)/(region area); we evaluate an easy split ($r\!<\!0.3$) and a hard split ($r\!\geq\!0.7$). 
We report the success rate (all pieces inside the target region without overlap) and the validity rate (non-overlap only).

\conciseparagraph{Result}
We report the results for the Rectangle composition in Table~\ref{tab:ch5_rectangle}. We evaluate the success rate, where a successful instance should have all the pieces in the target region without overlapping, and the validity rate for the non-overlapping constraints. Overall, the conclusions largely mirror the Tangram Assembly results; the notable difference is that several methods perform reasonably well on easy rectangle tasks due to the relaxed setting.
Auto-regressive and diffusion methods are competitive on the easy split ($51$--$97\%$ success) and then collapse on the hard split ($1$--$30\%$), a drop that neither stronger energy guidance nor adversarial PPO closes. GAG MCTS stays at $95$--$96\%$ easy and $50$--$53\%$ hard, widening the gap from a few points on easy to $\sim\!20$--$50$ points on hard. Section~\ref{sec:analysis} explains the shape of this easy-to-hard trajectory using the barrier bounds.

{
\subsection{Analysis: why the feasible-mass bound explains the observed pattern}
\label{sec:analysis}

The empirical pattern above is consistent with the feasible-set theory of Section~\ref{sec:theory}: wherever the relative feasible mass $\mu_y$ is small, continuous baselines fail. We now show that $\mu_y$ is small in our two tasks through qualitatively different mechanisms, and that the resulting regimes (shown in Table~\ref{tab:barrier_summary}) are consistent with the observed performance. Full derivations are in Appendix~\ref{app:theory_apply}.

\conciseparagraph{Tangram (contact precision)} The connectivity constraints confine the feasible set to a stratified union of submanifolds $\mathcal{F}_\mathrm{tan}=\bigcup_\alpha\mathcal{M}_\alpha$, indexed by contact signatures $\alpha$. Applying the precision-barrier tube-volume bound of~\citet{soh2026hallucination} per stratum and summing gives $\mathrm{Vol}(\mathcal{F}_\mathrm{tan}^\epsilon)\leq\sum_\alpha C_\alpha\,\epsilon^{d-k_\alpha}$, where $k_\alpha$ is the dimension of stratum $\alpha$ and $d=21$. PCA on a tangent-move walker gives an empirical relative-motion estimate of $\hat k\!\approx\!3.5$ (range $2$--$5$; Appendix~\ref{app:apply_tangram}), but this single-piece walker may miss joint articulations and is not used as an upper bound. Conservatively counting six spanning-tree contacts as each removing one scalar constraint gives $\mu_\mathrm{tan}(\epsilon)\lesssim C_\mathrm{tan}\,\epsilon^{6}$; at $\epsilon\!=\!0.01$ the volume factor is $10^{-12}$, so the corresponding required concentration factor is of order $10^{12}$.

\conciseparagraph{Rectangle (pose-volume depletion)} Without connectivity constraints, the feasible set can be full-dimensional when slack is ample, so contact-precision arguments do not apply. Instead, the relative feasible mass decomposes sequentially: $\mu_\mathrm{rect}(y)=\prod_{t=1}^{N}\bar\alpha_t(y)$, where $\bar\alpha_t$ is the average fraction of individually legal poses for piece $t$ that remain non-overlapping after a random valid prefix of length $t{-}1$ (Appendix~\ref{app:apply_rect}). On the easy split ($\bar{r}\!=\!0.22$), $k\!\approx\!d$ and ample free space suggests the $\bar\alpha_t$ factors remain moderate, so $\mu$ is not too small and diffusion performs well. On the hard split ($\bar{r}\!=\!0.76$), available pose volume collapses as the region fills, making $\mu_\mathrm{rect}$ exponentially small, explaining the sharp easy-to-hard degradation in Table~\ref{tab:ch5_rectangle}.

\conciseparagraph{Mapping to the method hierarchy} Guidance can increase the concentration factor $K_{\theta,\beta}$ but cannot change $\mu_y$, explaining why energy-guided diffusion improves non-overlap validity but fails to close the hard-split success gap. Autoregressive discretisation builds part of the contact structure into the action space, bypassing the global tube-hitting problem. PPO increases mass on actions that lead to successful completions. MCTS goes further by explicitly choosing actions that preserve future feasibility: in the rectangle decomposition, it can favour placements that preserve future feasibility rather than merely fitting immediately.

\begin{table}[htbp]
\centering\small
\caption{Feasible-mass regimes under the uniform reference measure, a task property independent of the generator. Tangram uses per-stratum PCA; rectangles use grid enumeration and RRT connectivity~\citep{lavalle1998rapidly} (Appendix~\ref{app:theory_apply}). Empirical results appear in Tables~\ref{tab:ch5_main} and~\ref{tab:ch5_rectangle}.}
\label{tab:barrier_summary}
\begin{tabular}{@{}lcccl@{}}
\toprule
Task / regime & $d$ & $\hat k$ & $\mu_y$ regime & Source of small $\mu_y$ \\
\midrule
Rectangle easy ($\bar{r}\!=\!0.22$)   & $6$  & $5$--$6$ & moderate              & ---     \\
Rectangle hard ($\bar{r}\!=\!0.76$)   & $12$ & $0$--$3$ & very small & pose-volume depletion \\
Tangram (single stratum, mean)  & $21$ & $\approx 3.5$\rlap{$^*$}    & very small & contact precision \\
\bottomrule
\multicolumn{5}{l}{\footnotesize $^*$Empirical relative-motion proxy; excludes $3$ global rigid-body DOFs.}
\end{tabular}
\end{table}
}

\section{Discussion}
\label{sec:discussion}

\conciseparagraph{Findings}
Across both tasks, continuous diffusion baselines perform well only in the one regime where $\mu_y$ is not small: easy rectangle, with $k\!\approx\!d$ and ample available pose volume. Wherever $\mu_y$ becomes small: whether through contact precision (tangram) or pose-volume depletion (hard rectangle), diffusion degrades substantially, consistent with the feasible-mass bound (Eq.~\ref{eq:feasible_mass_bound}). Discrete autoregressive and masked models partially bypass the tube-hitting problem and achieve higher validity, but still fall short when future completability matters. GAG MCTS is the only tested approach that stays near-saturation across all regimes. On this class of problems, whose feasible sets are low-dimensional or sequentially depleted, the evidence supports preferring discrete, constraint-aware action spaces with search over continuous density matching.

\conciseparagraph{Limitations and future work}
Our work has two main limitations. First, the CLIP-based reward for tangram is weakly calibrated and prone to overfitting under the limited KiloGram data, which motivates augmenting it with hybrid learned-plus-symbolic verifiers. Second, search has substantially higher inference cost than single-pass baselines (Appendix~\ref{app:inference_cost}); bringing the gap down through better simulation budgets, amortised value networks, or distilled policies is an obvious direction. Discrete diffusion remains a promising direction when it is defined over an on-manifold or sequential action space; a naive full-layout formulation is impractical for tangram because the joint pose space grows exponentially with the number of pieces. Looking forward, the most natural extension is to apply the on-manifold sequential recipe to other tasks in the same class: 3D CAD layout, floor plans, molecule generation, and engineering inverse design, where feasible sets are similarly structured. A longer-term open question is whether the benefits of search can be recovered within a continuous framework, for instance, through \emph{manifold-guided} samplers whose denoising targets a mixed discrete-continuous on-manifold action space rather than unconstrained coordinates.

{
    \small
    \bibliographystyle{plainnat}
    \bibliography{references}
}

\clearpage \appendix

\section{Feasible-Set Theory for Constrained Layout Generation}
\label{app:theory_apply}

This appendix develops the feasible-set framework summarised in Section~\ref{sec:theory} and applies it to our two tasks, building on ideas from~\citet{soh2026hallucination}. Section~\ref{app:theory_statements} states the general feasible-mass bound and its two specialisations; Sections~\ref{app:apply_rect} and~\ref{app:apply_tangram} measure the geometric quantities that determine $\mu_y$ for rectangle composition and tangram, and compare to the empirical diffusion-baseline validity.

\subsection{General feasible-mass framework}
\label{app:theory_statements}

Let $y$ denote a task instance (text prompt, piece set, target region). Let $\mathcal{X}_y$ be the continuous layout space; for planar rigid pieces a single pose is $(p,\theta)\in\mathbb{R}^2\times S^1$. Define a \emph{geometric} violation score $\mathrm{viol}_y(x)\geq 0$ measuring overlap and connectivity gap. The tube-volume bounds below apply to this geometric feasible set; semantic matching (e.g.\ text--silhouette alignment) is handled by the reward and can only shrink the feasible set further, making the bound conservative.

\begin{definition}[Tolerant feasible set]
For tolerance $\epsilon\geq 0$, $\mathcal{F}_y^\epsilon:=\{x\in\mathcal{X}_y:\mathrm{viol}_y(x)\leq\epsilon\}$. The exact feasible set is $\mathcal{F}_y:=\mathcal{F}_y^0$.
\end{definition}

\begin{definition}[Relative feasible mass]
Assuming $\mathcal{X}_y$ is bounded with uniform reference density $p_y$,
$\mu_y(\epsilon):=\mathrm{Vol}(\mathcal{F}_y^\epsilon)/\mathrm{Vol}(\mathcal{X}_y)$.
\end{definition}

Let a diffusion, flow, or other continuous generator induce a density $p_{\theta,\beta}(x\mid y)$, where $\theta$ are model parameters and $\beta$ an optional guidance scale. Define the concentration factor
\begin{equation}
K_{\theta,\beta}(y,\epsilon) \;:=\; \operatorname{ess\,sup}_{x\in\mathcal{F}_y^\epsilon}\frac{p_{\theta,\beta}(x\mid y)}{p_y(x)}.
\end{equation}

\begin{proposition}[Feasible-mass bound]
For any one-shot continuous generator,
\begin{equation}
S_{\theta,\beta}(y,\epsilon) \;:=\; \Pr\nolimits_{x\sim p_{\theta,\beta}}[x\in\mathcal{F}_y^\epsilon]
\;\leq\; K_{\theta,\beta}(y,\epsilon)\;\mu_y(\epsilon).
\label{eq:feasible_mass_bound_app}
\end{equation}
\end{proposition}
\begin{proof}
Since $p_y$ is uniform on $\mathcal{X}_y$ we have $p_y(x)=1/\mathrm{Vol}(\mathcal{X}_y)$, so
\begin{align}
S_{\theta,\beta}(y,\epsilon)
&= \int_{\mathcal{F}_y^\epsilon} p_{\theta,\beta}(x\mid y)\,dx
= \int_{\mathcal{F}_y^\epsilon} \frac{p_{\theta,\beta}(x\mid y)}{p_y(x)}\,p_y(x)\,dx \notag\\
&\leq \operatorname{ess\,sup}_{x\in\mathcal{F}_y^\epsilon}\frac{p_{\theta,\beta}(x\mid y)}{p_y(x)}
\int_{\mathcal{F}_y^\epsilon} p_y(x)\,dx
= K_{\theta,\beta}(y,\epsilon)\;\mu_y(\epsilon). \qedhere
\end{align}
\end{proof}

If $\mu_y(\epsilon)=10^{-n}$, the generator needs a concentration factor of order $10^n$ near the feasible set to achieve constant validity. Guidance changes $K_{\theta,\beta}$ but not $\mu_y$. The next two subsections show how $\mu_y$ becomes small in qualitatively different ways for our two tasks.

\paragraph{Relation to the barriers of~\citet{soh2026hallucination}.} The feasible-mass bound unifies and extends two barriers from~\citet{soh2026hallucination}. Their \emph{precision barrier} bounds the probability that a smooth decoder lands in the $\epsilon$-tube of a compact $C^1$ submanifold; this is a special case of our framework applied per stratum when $\mathcal{F}_y$ is a union of submanifolds. Their \emph{horizon barrier} bounds multi-step success as $\prod_t\gamma_t$ where $\gamma_t$ is the per-step probability of landing in the progress set; our available-pose-volume decomposition $\mu_\mathrm{rect}=\prod_t\bar\alpha_t$ is a concrete analogue of this product structure for rectangle packing under our uniform reference measure, though $\bar\alpha_t$ measures available pose volume rather than policy-dependent progress probability.

\subsection{Application to rectangle composition}
\label{app:apply_rect}

Rectangle feasibility requires every piece inside the target region and no pairwise overlap (strict inequalities only), with no connectivity requirement. The ambient space is $\mathbb{R}^{3N}$ for $N$ rectangles ($3$ DOFs per piece: centre and orientation). Formally, the feasible set is full-dimensional wherever there is slack ($k=3N$), so the precision-barrier submanifold framing of~\citet{soh2026hallucination} does not directly apply. Instead, the key quantity is how much pose volume remains available for each rectangle as the packing fills up.

\paragraph{Estimating local slack $(d,\hat k,M)$.} Table~\ref{tab:pca_rect} reports a local PCA slack estimate $\hat k$ (single-piece-perturbation sampler) together with the continuous mode count $M$ (number of path-connected components of the feasible manifold, estimated by the RRT connectivity method described below). Easy and hard configurations are matched to the dataset splits in Section~\ref{sec:exp_rect}. Although rectangle feasibility is topologically full-dimensional where slack exists, $\hat k$ decreases with $r$: the easy configuration has $\hat{k}\in\{5,6\}$ per packing, and the hard configuration has $\hat{k}\in\{0,3\}$ with most packings locally locked.

We estimate $M$ via a multi-tree RRT analysis. We first enumerate a representative set of valid integer-grid, axis-aligned packings by exhaustive grid search. Each enumerated packing seeds an independent RRT that grows by steering toward random feasible perturbations with a step size scaled to the fill ratio. When two distinct trees come within a fixed connection radius (in $\ell_2$ configuration distance), their roots are merged by union-find, reducing the component count. After a fixed node budget per tree, the number of surviving union-find components gives $\hat M$. The estimate is upper-biased: components that are truly connected but never bridged within the RRT budget are counted as separate, so $\hat M$ may overcount. On the easy split the ample free space means any packing can quickly slide to any other, collapsing all trees to one component ($M\!=\!1$). On the hard split no piece can move without immediate overlap, so trees cannot bridge across packings and $\hat M$ matches the number of seed packings ($M\!\sim\!10^3$).

\begin{table}[htbp]
\centering\small
\caption{Local PCA slack estimate $\hat{k}$ and continuous mode count $M$ for rectangle-packing configurations at two fill ratios $r$. $M$ is the estimated number of path-connected components of the feasible manifold, obtained by the multi-tree RRT connectivity method described above. PCA is run on a random subset of discrete packings per configuration with single-piece perturbations; $\hat k$ should not be read as the formal topological dimension, which is full-dimensional wherever strict-inequality slack exists.}
\label{tab:pca_rect}
\begin{tabular}{@{}lccccl@{}}
\toprule
Config & Average fill $\bar{r}$  & $d=3N$ & $M$ (continuous) & $\hat{k}$  \\
\midrule
easy     & $0.22$ & $6$ & $\mathbf{1}$    & $5$--$6$ \\
hard     & $0.76$ & $12$ & $\mathbf{\sim\!10^3}$ & $0$--$3$ \\
\bottomrule
\end{tabular}
\end{table}

\paragraph{Available-pose-volume decomposition.} Let $R_y\subset\mathbb{R}^2$ be the target region for instance $y$. For rectangle $i$ with shape $Q_i$, define the available pose set $D_i(F):=\{(c,\theta):c+Q_i(\theta)\subseteq F\}$ for any free region $F\subseteq R_y$, and its volume $V_i(F):=\mathrm{Vol}(D_i(F))$. Let $P_i(x_i)=c_i+Q_i(\theta_i)$ denote the placed rectangle. We choose the reference layout space $\mathcal{X}_y=\prod_{i=1}^N D_i(R_y)$ (each rectangle independently uniform among poses fitting the empty region). The only remaining hard constraint is pairwise non-overlap, so
\begin{equation}
\mu_\mathrm{rect}(y) \;=\; \Pr\nolimits_{x\sim\mathrm{Unif}(\mathcal{X}_y)}\!\bigl[P_i(x_i)\cap P_j(x_j)=\emptyset\;\;\forall\,i\neq j\bigr].
\label{eq:rect_mu_def}
\end{equation}

\begin{proof}[Derivation of the product decomposition]
Fix an ordering $i_1,\dots,i_N$ and let $E_t$ be the event that the first $t$ rectangles in this order are pairwise non-overlapping ($E_0$ is trivially true). By the chain rule of conditional probability,
\begin{equation}
\mu_\mathrm{rect}(y) = \Pr[E_N] = \prod_{t=1}^{N}\Pr[E_t\mid E_{t-1}].
\label{eq:rect_chain}
\end{equation}
Given a valid prefix of length $t{-}1$, the remaining free region is $F_{t-1}=R_y\setminus\bigcup_{s=1}^{t-1}P_{i_s}(x_{i_s})$. Under the reference measure each $x_{i_t}$ is uniform on $D_{i_t}(R_y)$, so the probability that rectangle $i_t$ avoids all previously placed rectangles (i.e.\ that $P_{i_t}(x_{i_t})\subseteq F_{t-1}$) is $V_{i_t}(F_{t-1})/V_{i_t}(R_y)$. Taking the expectation over valid prefixes,
\begin{equation}
\Pr[E_t\mid E_{t-1}] = \mathbb{E}\!\left[\frac{V_{i_t}(F_{t-1})}{V_{i_t}(R_y)}\;\middle|\;E_{t-1}\right] =: \bar\alpha_t(y).
\end{equation}
Substituting into~\eqref{eq:rect_chain} gives the decomposition
\begin{equation}
\mu_\mathrm{rect}(y) \;=\; \prod_{t=1}^{N}\bar\alpha_t(y).
\label{eq:rect_decomp}
\end{equation}
\end{proof}

The factor $\bar\alpha_t$ has a direct physical meaning: it is the average fraction of individually legal poses for rectangle $i_t$ that remain legal after a random valid prefix of length $t{-}1$.

\paragraph{Connectivity of the feasible set.} RRT sampling (Table~\ref{tab:pca_rect}) shows $M\!=\!1$ on the easy split but $M\!\sim\!10^3$ disconnected modes on the hard split, indicating that feasible-set disconnection may compound the pose-volume-depletion effect in the hard regime.

\paragraph{Mapping to observed performance.} On the easy split, $\mu_\mathrm{rect}$ is not too small, so diffusion only needs moderate concentration $K_{\theta,\beta}$ and can perform well ($89$--$97\%$ success; Table~\ref{tab:ch5_rectangle}). On the hard split, the $\bar\alpha_t$ factors collapse and a diffusion sampler must provide a much larger concentration factor. Energy guidance increases $K_{\theta,\beta}$ by pushing samples away from overlaps, but cannot change $\mu_\mathrm{rect}$, explaining why stronger guidance improves non-overlap validity but fails to close the hard-split success gap. The masked autoregressive model samples from $D_{i_t}(F_{t-1})$ at each step, removing many immediate failures, but local feasibility does not guarantee future completability. MCTS uses lookahead to favour actions that preserve future feasibility, explaining its advantage over PPO and autoregressive baselines.

\subsection{Application to tangram assembly}
\label{app:apply_tangram}

A tangram configuration of the seven pieces lives in ambient $\mathcal{X}\subset\mathbb{R}^{21}$ with coordinates $(x_i,y_i,\theta_i)_{i=1}^7$ (flip is discrete and ignored). Feasibility is \emph{(i) global connectivity}: the contact graph over all seven pieces is connected (each piece shares at least one edge or vertex with another such that no subset is isolated), and \emph{(ii) pairwise non-overlap}. Non-overlap is an open inequality; connectivity via shared contacts is an active equality constraint that pins configurations onto a lower-dimensional set. Sharing an \emph{edge} (rather than a point) still permits sliding: two pieces sharing a segment can translate tangentially along that segment while remaining connected and non-overlapping, and such slides compose across the contact graph.

\paragraph{Counting $(d,k)$ from the contact graph.} Let $G=(V,E)$ be the contact graph of a valid configuration, $|V|=7$, with an edge per pair of pieces in direct contact. Each contact imposes equality constraints whose multiplicity depends on type: \emph{edge--edge} contacts remove $2$ DOFs (normal separation and relative planar orientation); \emph{vertex--edge} contacts pin the vertex onto a line, removing $1$ DOF while leaving tangential sliding and rotation free; \emph{vertex--vertex} contacts pin two coordinates (the vertices coincide), removing $2$ DOFs but leaving rotation about the shared point free. For a spanning tree with $|E|=6$ generic edge--edge contacts,
\begin{equation}
k \;\leq\; 21 \;-\; \sum_{e\in E}\mathrm{mult}(e) \;=\; 21 - 12 \;=\; 9,
\label{eq:tangram_k_theory}
\end{equation}
which includes $3$ global rigid-motion DOFs. Additional cycle edges can only decrease the dimension or leave it unchanged; redundancy means they may decrease it less than a naive subtraction of all edge multiplicities would suggest. Mixes of contact types also change the per-contact multiplicity. The upper bound $k\leq 9$ corresponds to a spanning tree of pure edge--edge contacts; conservatively counting each of the six spanning-tree contacts as removing at least one scalar constraint gives $k\leq 15$, which still yields the $\mu_\mathrm{tan}\lesssim C_\mathrm{tan}\,\epsilon^{6}$ bound used below.

\paragraph{Empirical $(d,k,M)$ via a sequential random walker.} We probe the local dimensionality of the feasible set with a sequential random tangent-move walker. Starting from a valid configuration $c$, each sample is produced by the following \emph{sweep}: (i) draw $K\sim\mathrm{Uniform}\{1,\dots,7\}$; (ii) for each of $K$ iterations, pick a piece $i$ uniformly at random, pick one of its current edge/vertex contacts uniformly at random, and propose a tangent move (slide along a shared edge by $s\sim\mathcal{N}(0,\sigma)$, or rotate about a shared vertex by $\alpha\sim\mathcal{N}(0,\sigma/r)$ where $r$ is the piece's characteristic radius); (iii) accept the move iff the full $21$-dim configuration is still feasible (no overlaps and all pieces connected), otherwise skip that piece. The resulting configuration is recorded as one sample and the walker continues from it. PCA on the accepted-sample deviations (relative to the start, in $\mathbb{R}^{21}$) reports $\hat{k}$ as the number of principal components explaining $\geq 99\%$ of variance. A \emph{single-stratum} variant rejects any sample whose contact-graph signature differs from the start's, isolating the local tangent dimension of one stratum.

\emph{Caveat.} The walker samples contact-preserving \emph{relative} moves among pieces and excludes global rigid-body translations or rotations. The $3$ global DOFs therefore contribute little variance and are largely absent from $\hat k$. Because the proposals move one piece at a time and reject contact-signature changes in the single-stratum variant, the PCA estimates provide empirical lower-bound proxies for relative local motion; joint multi-piece articulations and global pose can add dimensions. The conservative $\epsilon^6$ bound is unchanged, since it counts only one scalar constraint per spanning-tree contact independently of the PCA reading.

We run the walker on $6$ valid 7-piece configurations. To obtain a diverse set, we start from a hand-designed \emph{chain7} configuration (two large triangles connected indirectly through a chain of edge and point contacts) and take $6$ widely-separated snapshots of the walker itself at a coarser budget; each snapshot has a distinct contact signature from all others. Per-configuration results are in Table~\ref{tab:pca_7piece_multi} and the aggregate in Table~\ref{tab:pca_7piece}. Averaging over configurations, $\hat k=3.67\pm 1.03$ under the union rule and $\hat k=3.50\pm 1.05$ under the single-stratum rule. Since the walker omits global rigid motion and may miss joint articulations, these values are empirical local-motion estimates rather than formal manifold dimensions; the conservative contact count above gives the dimension bound used in the mass estimate. Every configuration has signature-component count $1$ under the union rule, confirming that the visited signatures are feasibly connected rather than disjoint modes.

\paragraph{Connectivity of the feasible set.} All $6$ sampled configurations collapse to a single connected component under the union-find walker (Tables~\ref{tab:pca_7piece}--\ref{tab:pca_7piece_multi}). Among the configurations we tested, the only source of genuine disconnection is parallelogram chirality ($M\leq 2$). This suggests that the tangram failure of diffusion baselines is attributable to small feasible mass from contact precision rather than to disconnection of the feasible set; configurations outside our sample may contain additional disconnected components.

\begin{table}[htbp]
\centering\small
\caption{PCA-based \emph{relative-motion} estimate for tangram (ambient $d=21$), averaged over $6$ valid 7-piece configurations. The walker samples contact-preserving relative moves and does not inject global translations or rotations, so $\hat k$ excludes the $3$ global rigid-body DOFs and may miss joint multi-piece articulations. Produced by the sequential random tangent-move walker ($\sigma=0.005$, $3000$ samples per configuration). \emph{Single-stratum} rejects samples that change the contact-graph signature; \emph{union} accepts any feasible sample. Per-configuration breakdown in Table~\ref{tab:pca_7piece_multi}.}
\label{tab:pca_7piece}
\begin{tabular}{@{}lcc@{}}
\toprule
Rule & mean $\hat{k}$ (99\%) & $\hat k$ distribution over $6$ configs \\
\midrule
single-stratum & $3.50\pm 1.05$ & $\{2, 3, 3, 4, 4, 5\}$ \\
union          & $3.67\pm 1.03$ & $\{2, 3, 4, 4, 4, 5\}$ \\
\bottomrule
\end{tabular}
\end{table}

\begin{table}[htbp]
\centering\small
\caption{Per-configuration PCA results for the $6$ valid tangram configurations used in Table~\ref{tab:pca_7piece}. ``\# sig.'' is the number of distinct contact signatures visited and ``\# comp.'' is the number of connected components in the signature-adjacency graph; \# comp.\ $=1$ means the visited signatures are feasibly connected and therefore \emph{not} disjoint topological modes.}
\label{tab:pca_7piece_multi}
\begin{tabular}{@{}lccc|ccc@{}}
\toprule
 & \multicolumn{3}{c|}{Union} & \multicolumn{3}{c}{Single-stratum} \\
Config & $\hat{k}$ & \# sig. & \# comp. & $\hat{k}$ & move accept & top-3 SVs \\
\midrule
chain7\_00 & $4$ & $27$ & $1$ & $3$ & $74.8\%$ & $23.2, 3.1, 2.0$ \\
chain7\_01 & $5$ & $44$ & $1$ & $5$ & $73.4\%$ & $8.9, 5.1, 2.8$ \\
chain7\_02 & $4$ & $47$ & $1$ & $4$ & $64.0\%$ & $15.3, 4.0, 3.4$ \\
chain7\_03 & $2$ & $16$ & $1$ & $4$ & $68.7\%$ & $11.1, 5.0, 3.5$ \\
chain7\_04 & $4$ & $29$ & $1$ & $3$ & $61.1\%$ & $21.9, 7.7, 3.9$ \\
chain7\_05 & $3$ & $30$ & $1$ & $2$ & $63.6\%$ & $16.1, 2.6, 1.4$ \\
\bottomrule
\end{tabular}
\end{table}

\paragraph{Feasible-mass bound on tangram.} The tangram feasible set is a union of submanifolds $\mathcal{F}_\mathrm{tan}=\bigcup_\alpha\mathcal{M}_\alpha$ indexed by contact signatures. Applying the tube-volume bound of~\citet{soh2026hallucination} per stratum and summing,
\begin{equation}
\mathrm{Vol}(\mathcal{F}_\mathrm{tan}^\epsilon) \;\leq\; \sum_{\alpha\in\mathcal{A}}\mathrm{Vol}(\mathcal{M}_\alpha^\epsilon) \;\leq\; \sum_{\alpha\in\mathcal{A}} C_\alpha\,\epsilon^{d-k_\alpha},
\label{eq:tangram_tube}
\end{equation}
where $\mathcal{M}_\alpha^\epsilon$ is the $\epsilon$-tube around stratum $\alpha$. The PCA walker estimates a relative local-motion dimension $\hat k\approx 3.5$; the conservative bound below relies on contact counting instead, since the single-piece proposal may miss joint articulations. Even conservatively counting each of the six spanning-tree contacts as removing only one DOF gives $\mu_\mathrm{tan}(\epsilon)\lesssim C_\mathrm{tan}\,\epsilon^6$. At $\epsilon=0.01$, the volume factor is $\epsilon^6=10^{-12}$ (up to the unknown constant $C_\mathrm{tan}$), so the corresponding required concentration factor is of order $\epsilon^{-6}=10^{12}$. These figures are order-of-magnitude illustrations---the bounds are loose because $C_\alpha$ and the exact tolerance are unspecified---but they qualitatively indicate that the required concentration is far beyond what current diffusion samplers are observed to provide, consistent with the $0$--$4\%$ validity of energy-guided diffusion on tangram in Table~\ref{tab:ch5_main}.

\paragraph{Mapping to observed performance.} Diffusion samples the whole layout in continuous coordinates and must solve the global tube-hitting problem in $d=21$; guidance can improve $K_{\theta,\beta}$ but cannot change $\mu_\mathrm{tan}$. The autoregressive model discretises placement by choosing from anchor-aligned actions, building part of the contact structure into the action parameterisation and bypassing the tube-hitting problem, explaining why plain autoregressive decoding already reaches much higher validity ($54$--$88\%$). PPO further increases mass on progress actions (those leading to successful final assemblies). MCTS adds lookahead: many locally valid placements leave poor future completions or poor semantic matches, and search estimates future value to amplify conditionally completable, high-reward continuations.

\subsection{What the feasible-mass bound implies for our experiments}
\label{app:theory_implications}

\begin{itemize}
    \item \emph{Mode collapse vs.\ invalidity tradeoff on rectangle.} On the hard rectangle split, feasible-set disconnection ($M\!\sim\!10^3$) may further reduce $K_{\theta,\beta}$ for generators that try to cover all modes. PPO without adversarial refinement may concentrate on a few modes, improving validity but potentially reducing mode coverage (not directly measured in Table~\ref{tab:ch5_rectangle}).
    \item \emph{Tightening constraints hurts diffusion disproportionately.} As constraints tighten, $\mu_y$ shrinks through $\epsilon^{d-k}$ for tangram or through collapsing $\bar\alpha_t$ factors for rectangles. This is visible in Table~\ref{tab:ch5_rectangle}: diffusion and PPO are competitive on the easy split but collapse on the hard split where $\mu_\mathrm{rect}$ is exponentially smaller.
    \item \emph{Why search is well-matched.} MCTS operates on a \emph{discrete} action space where the tube-density requirement is vacuous. For the sequential decomposition $\mu=\prod_t\bar\alpha_t$, search is designed to favour actions that preserve future feasibility (i.e.\ keep future $\bar\alpha_t$ from collapsing), rather than merely fitting immediately. Tree search with a learned policy-value prior and a reward-based verifier directly addresses both the contact-precision and pose-volume-depletion sources of small $\mu_y$.
    \item \emph{Caveat (energy guidance).} Guidance gradients increase $K_{\theta,\beta}$ by steering samples toward low-violation regions, but cannot change $\mu_y$. Our experiments include guided diffusion baselines with varying $\beta$ and still observe sharp degradation on the hard split, consistent with the view that guidance mitigates but does not eliminate the structural issue.
\end{itemize}

\section{Preliminary: Gumbel MCTS}\label{app:gumbel_muzero}

Gumbel MuZero~\citep{danihelka2022policy} modifies AlphaZero-style search~\citep{silver2017mastering} by using Gumbel-Top-$k$ sampling and sequential halving at the root. Our implementation uses this Gumbel root-selection operator, while advancing tree transitions with the exact geometric simulator rather than a learned latent dynamics model. We therefore refer to the operator used in this paper as Gumbel MCTS.

\paragraph{Gumbel-Top-k trick.} Gumbel MCTS uses the Gumbel-Max trick \citep{gumbel1954statistical, yellott1977relationship, jang2016categorical, kool2019stochastic} to sample $n$ actions without replacement from policy logits:
\begin{align}
    (g\in\mathbb{R}^k) &\sim \mathrm{Gumbel}(0)\\
    A_1 &= \underset{a}{\arg\max}(g(a) + \ell_\theta(s,a))\\
    &\cdots\notag\\
    A_n &= \underset{a\notin \{A_1, \cdots, A_{n-1}\}}{\arg\max}(g(a) + \ell_\theta(s,a)).
\end{align}
Here $\ell_\theta(s,a)$ denotes the unnormalised policy logit, and we use $\texttt{argtop}(g + \ell_\theta, n) = \{A_1, A_2, \cdots, A_n\}$ for the Gumbel-Top-k samples. 

\paragraph{Sequential Halving with Gumbel.} Sequential Halving \citep{karnin2013almost} is a bandit algorithm for simple regret minimisation. It aims at maximising the $Q$-value of the last or the finally selected action in simulations, which differs from the cumulative regrets that maximise the $Q$ from all $n$ simulations \citep{tolpin2012mcts}. 

\begin{algorithm}[htbp]
\caption{Sequential Halving with Gumbel}
    \begin{algorithmic}[1]
    \Procedure{SeqHalGumbel}{$s, N, K, g, \ell_\theta, f, \sigma$}\Comment{$\ell_\theta$: policy logits; $f(s,a,m)$: average return from $m$ simulations forced to start with action $a$.}
    \State $\texttt{Seq} \leftarrow \texttt{argtop}(g + \ell_\theta(s,\cdot), K)$
    \While{$|\texttt{Seq}|>1$}
        \For{$a \in \texttt{Seq}$}
            \State $q(s, a) \leftarrow f(s, a, \lfloor\frac{N}{|\texttt{Seq}|\times\log_2(K)}\rfloor)$ 
        \EndFor
        \State $\texttt{Seq}\leftarrow$ set of $\lceil \frac{|\texttt{Seq}|}{2}\rceil$ actions in $\texttt{Seq}$ with the largest $(g(a)+\ell_\theta(s,a)+\sigma(q(s,a)))$
    \EndWhile 
    \State \textbf{return} the remaining action in $\texttt{Seq}$
    \EndProcedure
  \end{algorithmic}
 \label{alg:SeqHalving}
\end{algorithm}

Algorithm~\ref{alg:SeqHalving} shows Sequential Halving with Gumbel. The original Sequential Halving samples actions uniformly at random, while the Gumbel variant samples from learned policy logits without replacement. It then simulates those sampled actions for $\lfloor\frac{N}{|\texttt{Seq}|\times\log_2(K)}\rfloor$ trials each, keeps the half with the highest transformed average return, and repeats until one action remains.

\paragraph{Gumbel MCTS}
Using Sequential Halving with Gumbel, we modify the root action selection procedure in Monte Carlo tree search. The pseudocode is shown in Alg~\ref{alg:gumbelMuzeroMCTS}. 

\begin{algorithm}[htbp]
\caption{Gumbel MCTS with exact simulator transitions}
    \begin{algorithmic}[1]
    \Procedure{Search}{$s, N, K, g, \ell_\theta, \sigma, \tau$} \Comment{$s$: root state; $\tau$: tree; $N$: simulation budget.}
    \State Define $f(s,a,m)$ as the average of $m$ calls to \Call{Simulate}{$s, a, \tau, \texttt{False}, 0, \texttt{True}$}
    \State\Return \textsc{SeqHalGumbel($s, N, K, g, \ell_\theta, f, \sigma$)}
    \EndProcedure

\Procedure{Simulate}{$s, a, \tau, \texttt{done}, \texttt{depth}, \texttt{forced}$}
    \If{$\gamma^\texttt{depth}<\epsilon$ or $\texttt{done}=\texttt{True}$ }
        \State \textbf{return} 0
    \EndIf
    \If{$s$ is a leaf node and $\texttt{forced}=\texttt{False}$}
        \State Update $\tau$ by adding the new state node
        \State $\forall a\in A, N(s, a)\leftarrow 0, q(s, a)\leftarrow 0$
        \State \textbf{return} $v_\theta(s)$ 
    \EndIf
    \If{$\texttt{forced}=\texttt{True}$}
        \State $a^* \leftarrow a$ \Comment{Explicitly evaluate the root action selected by sequential halving}
    \Else
        \State $p \leftarrow \mathrm{softmax}(\ell_\theta(s,\cdot))$
        \State $a^*\leftarrow\arg\max_{b\in A(s)} \left(q(s,b)+c_\mathrm{puct}p(b)\frac{\sqrt{N(s)}}{1+N(s,b)}\right)$
    \EndIf
    \State Simulate action $a^*$ from node $s$ to get next state node $s'$, reward $r$, and \texttt{done}. 
    \State $R \leftarrow r + \gamma\cdot$\Call{Simulate}{$s', a^*, \tau, \texttt{done}, \texttt{depth + 1}, \texttt{False}$}
    \State $N(s, a^*) \leftarrow N(s, a^*) + 1$, $N(s)\leftarrow N(s) + 1$
    \State $q(s, a^*)\leftarrow q(s, a^*) + \frac{R-q(s, a^*)}{N(s, a^*)}$
    \State \textbf{return} $R$
\EndProcedure
  \end{algorithmic}
 \label{alg:gumbelMuzeroMCTS}
\end{algorithm}

Gumbel MCTS uses the search procedure to generate planning data, which is then used to train the policy and value heads. The root-selection procedure has the policy-improvement guarantee from the Gumbel MuZero analysis, i.e., 
\begin{align}
    &q(s, \underset{a\in\texttt{argtop}(g, \pi, k)}{\arg\max}(g(a) + \pi(s, a) + \sigma(\bar{q}(s, a)))\geq \notag \\& q(s, \underset{a\in\texttt{argtop}(g, \pi, k)}{\arg\max}(g(a) + \pi(s, a))).
\end{align}
where the $\bar{q}$ is estimated by MCTS and $\sigma$ is a transform function, such as the formula we use below:
\begin{equation}
    \sigma(q(a)) = (c_\mathrm{visit} + \max_b N (b))c_\mathrm{scale}q(a). 
\end{equation}
The equation is equivalent to $\mathbb{E}_{a\sim \mathrm{MCTS}}[q(s, a)] \geq \mathbb{E}_{a\sim\pi}[q(s, a)]$ according to the Gumbel-Max trick. Thus, the Gumbel root search can improve the policy even with a minimal number of MCTS simulations.

\section{Action masks for Tangram assembly}\label{app:action_mask}

We call the mask \emph{pairwise} because it checks feasibility only between the piece being placed and each piece already on the board, one pair at a time. It does not reason about three-way or higher-order interactions among placed pieces, nor does it verify that the remaining unplaced pieces can still complete a valid assembly. A placement that passes the pairwise mask is locally non-overlapping and anchor-aligned, but may still lead to a dead end when future pieces no longer fit.

The mask is implemented in two stages. In the \emph{precomputation} stage, valid actions are exhaustively generated by iterating through all piece combinations, transformations, and anchor-point alignments. For each pair of pieces (moved and reference), the algorithm considers eight rotation states ($0^\circ$ to $315^\circ$ in $45^\circ$ increments), up to 2 flip states (depending on piece symmetry), and anchor points derived from vertices and edge midpoints. Each candidate action is validated via polygon intersection tests, and duplicates are removed by relative-configuration hashing. The resulting valid-action list is stored as a JSON file and loaded at runtime.

During gameplay, the mask is \emph{dynamically} refined by applying contextual checks to the precomputed list: (1) the moved piece must be unplaced while the reference piece must already be placed, (2) flip and rotation IDs are within valid ranges for each piece type, (3) anchor-point indices are appropriate for the piece geometry (6 anchors for triangles, 8 for quadrilaterals), and (4) the reference piece's flip state matches the current environment state. This two-stage approach balances computational efficiency (precomputed geometric constraints) with runtime flexibility (state-dependent filtering).

This combinatorial growth is the main obstacle to a direct full-state discrete diffusion formulation. A discrete diffusion model over complete layouts would need to denoise over all ordered assignments of seven piece identities and their discrete placements. Even using a conservative $100$ locally possible placements per non-root piece gives $7!\times100^6\approx5\times10^{15}$ ordered assemblies. Counting raw anchor, rotation, and flip choices from the precomputation stage gives an upper-envelope count above $10^{24}$ before duplicate removal or feasibility filtering. Since most such joint states violate non-overlap or connectivity, a tractable discrete diffusion variant would need additional sequential or on-manifold factorisation rather than a direct full-layout state space.

When a locally valid placement later leads to an invalid terminal state or an uncompletable partial assembly, a negative penalty is applied so that the search algorithm learns to avoid such dead ends.

\section{Rectangle Packing Dataset Generation}\label{app:ch5_rectangle_gen}

\paragraph{Piece Definitions}
Three types of rectangular pieces are used: \texttt{1x2}, \texttt{2x3}, and \texttt{1x5}, each with allowed $0^\circ$ and $90^\circ$ rotations. The pieces are defined by their grid cell coverage.


\paragraph{Tiling and Packing Generation}
A recursive backtracking search is used to find a complete tiling of the target region using the available inventory. To create a packing problem, a random subset of the pieces from the full tiling is selected, with the fraction to keep sampled uniformly from $[0.1, 1.0]$.

\paragraph{Configuration Construction}
For each configuration, the target region dimensions, the set of pieces to pack, and the solution placements (including piece type, rotation, anchor position, and computed vertices) are recorded. All piece placements are centred by applying a global offset so that the region is centred at $(0,0)$.

\paragraph{Uniqueness and Output}
Each configuration is assigned a unique signature based on the region and piece set to ensure dataset diversity. Configurations are saved in JSON format for training, validation, and test splits. Example images are also generated for visualisation.

\section{Implementation details}
\label{app:ch5_implementation_details}

\subsection{Inference cost}
\label{app:inference_cost}

Table~\ref{tab:run-time} reports the average wall-clock inference cost per tangram configuration for GAG MCTS, energy-guided diffusion, and the autoregressive transformer, measured on the same hardware. The machine uses an Intel Xeon Platinum 8580 CPU with $192$ CPU cores available and $2$ NVIDIA A100 PCIe GPUs with $80$GB memory per GPU. GAG MCTS uses $200$ MCTS simulations per step; the diffusion baseline uses $1000$ denoising iterations. Search is roughly two orders of magnitude slower than a single-pass autoregressive decoder and about $2\times$ faster than the diffusion baseline at this simulation budget. The cost scales linearly in the number of simulations and could be reduced by pruning the simulation budget; we did not attempt such optimisation here.

\begin{table}[htbp]
\centering\small
\setlength\tabcolsep{4pt}
\caption{Inference cost on Tangram Assembly. GAG MCTS uses $200$ MCTS simulations per step; the diffusion baseline uses $1000$ denoising iterations. Average $\pm$ standard deviation over the test set, on the same hardware.}
\label{tab:run-time}
\begin{tabular}{@{}lccc@{}}
\toprule
                     & GAG MCTS        & Diffusion       & Auto-Reg \\ \midrule
Avg runtime (sec)    & $123.7\pm 15.4$ & $233.0\pm 20.1$ & $1.25\pm 1.3$ \\
\bottomrule
\end{tabular}
\end{table}

\subsection{Policy-value network: design rationale}
\label{app:policy_value_rationale}

A few choices deserve expansion beyond what is stated in Section~\ref{sec:method}. First, KiloGram has only a few thousand labelled configurations, so the policy-value network does not need, and would overfit with, a large pre-trained language backbone such as Llama; a pre-trained ViT and a frozen BERT are sufficient. Our architecture is inspired by OpenVLA, which also uses ViT as the image encoder and fine-tunes it during training; in our model, a frozen BERT extracts semantic features and a separate small transformer fuses them with the image features. Keeping BERT frozen is again a data-scarcity choice: it keeps the text-feature space stable and prevents the text encoder from co-adapting to the limited training set. Once trained, the policy network acts as a search heuristic inside MCTS: it prunes the tree by down-weighting unlikely actions and focusing expansion on promising continuations informed by previous search experience; the value network provides a rollout estimate without environment rollouts.

\subsection{Contrastive loss for reward training}
\label{app:contrastive_loss}

To avoid catastrophic forgetting of the pre-trained CLIP alignment during adversarial reward refinement (Section~4.2), we keep the standard bidirectional CLIP InfoNCE loss in the objective. For a batch of $N$ image--caption pairs sampled from the positive dataset $D_+$, let $v_i$ be the CLIP image feature of the $i$-th image and $w_i$ the CLIP text feature of the $i$-th caption. With temperature $T$, the bidirectional contrastive loss is
\begin{align}
    \mathcal{L}_\mathrm{cont}
    = -\frac{1}{N}\sum_i \log \frac{\exp(v_i\!\cdot\! w_i/T)}{\sum_j \exp(v_i\!\cdot\! w_j/T)}
      -\frac{1}{N}\sum_j \log \frac{\exp(v_j\!\cdot\! w_j/T)}{\sum_i \exp(v_i\!\cdot\! w_j/T)}.
\end{align}
The first term is an image-to-text cross-entropy (each image should be closer to its caption than to any other caption in the batch); the second term is the symmetric text-to-image cross-entropy. Together they recover the original CLIP training signal, which is combined with the adversarial preference loss $\mathcal{L}_\mathrm{pref}$ as $\mathcal{L}_\mathrm{pref} + \lambda\mathcal{L}_\mathrm{cont}$. In our experiments we use the temperature learned by the original CLIP checkpoint and $\lambda=1$.

\subsection{GAG MCTS implementation}

\paragraph{Framework:} The Tangram Assembly GAG MCTS agent is implemented in JAX/Flax, leveraging distributed training across multiple devices. The environment simulates the sequential assembly of tangram pieces with the exact geometric transition function, with each state represented as an image and associated instruction tokens.
\paragraph{Model Architecture:}
\begin{itemize}
\item \textit{Policy/Value Network:} The policy-value network uses a ViT image encoder and a frozen BERT text encoder, followed by a small transformer that fuses visual and language features. Two output heads produce action logits and value estimates.
\item \textit{Reward Model:} A CLIP-based~\citep{radford2021learning} reward model computes the alignment between the assembled image and the instruction, using the contrastive loss and the binary adversarial loss described in Section~\ref{sec:adversarial}.
\end{itemize}
\paragraph{Self-Play and Search:}
\begin{itemize}
\item \textit{Self-Play:} The agent generates training data via self-play, using Gumbel MCTS for action selection. The search uses the exact environment transition function and the learned policy-value network for priors and leaf evaluation.
\item \textit{Batching:} Self-play and training are parallelised across devices, with large batch sizes (e.g., 2048 for self-play).
\end{itemize}
\paragraph{Training:}
\begin{itemize}
\item \textit{Policy/Value Training:} The policy and value networks are trained using cross-entropy and mean squared error losses, respectively, with targets computed from self-play trajectories.
\item \textit{Reward Model Training:} The reward model is updated using both contrastive and binary adversarial losses, encouraging correct alignment between images and instructions while rejecting MCTS-generated near misses.
\item \textit{Optimisation :} Adam optimisers are used for both networks, with learning rates $5 \times 10^{-5}$ (policy/value) and $1 \times 10^{-8}$ (reward).
\item \textit{Checkpointing:} Models and optimizers are periodically checkpointed for recovery and evaluation.
\end{itemize}

\subsection{Diffusion model: Rectangle Packing}

\paragraph{Architecture:} We employ a Transformer-based diffusion model for rectangle assembly. Each rectangle is represented by its state $(x, y, \theta)$ and shape $(w, h)$, with a global target shape as an additional condition. States and shapes are embedded via linear layers and combined, while the target shape is provided to the Transformer decoder as a separate memory token. The model predicts denoising updates for all rectangles in parallel.
\paragraph{Classifier Guidance:} The sampling process incorporates differentiable guidance losses:
\begin{itemize}
\item \textit{Angle Loss:} Encourages $\theta$ to be close to multiples of $90^\circ$.
\item \textit{Overlap Loss:} Penalises overlapping rectangles using a differentiable centre-distance surrogate.
\item \textit{Region Loss:} Penalises rectangles placed outside a specified bounding region.
\end{itemize}
\paragraph{Sampling:} Diffusion sampling iteratively updates rectangle states using the model's prediction and the gradient of the guidance loss. Typical parameters include $100$ steps, a step size of $0.05$, and a guidance scale of $1.0$.

\subsection{Diffusion model: Tangram Assembly}

\paragraph{Architecture:} For tangram assembly, we utilise a Transformer encoder-based diffusion model. Each piece is represented by $(x, y, \text{rotation}, \text{flip}, \text{index})$. CLIP text embeddings of the target description are projected and concatenated with piece embeddings before being processed by the Transformer. The model predicts noise for $(x, y, \text{rotation})$ only; flip and index are fixed.
\paragraph{Dataset:} The dataset consists of tangram configurations paired with natural language descriptions. Each sample includes piece positions, types, and a binary target mask.
\paragraph{Guidance:} The guidance loss combines:
\begin{itemize}
\item \textit{Overlap Loss:} Penalises overlapping polygons using signed distance functions.
\item \textit{Angle Loss:} Encourages rotations to be multiples of $45^\circ$.
\end{itemize}
\paragraph{Diffusion Process:} We use a Denoising Diffusion Probabilistic Model (DDPM) scheduler with $1000$ timesteps. During training, the model is optimised to predict the added noise. During sampling, energy-function guidance~\citep{yu2023freedom} is optionally applied by backpropagating the energy through the denoising step.
\paragraph{Training Details:} Optimisation is performed using AdamW with a learning rate of $1 \times 10^{-4}$ and cosine annealing. The batch size is $512$. Validation is conducted periodically with visualisation and checkpointing. 

\subsection{Energy Function for Diffusion Guidance}

The energy function is designed to provide smooth, differentiable constraints during diffusion-based shape assembly, following~\citep{yu2023freedom}. It combines multiple loss terms, such as angle alignment, overlap penalty, and region constraints, into a single scalar objective. The general form, used for the rectangle baseline, is:
\begin{align}
f_{\text{guidance}}(\mathbf{s}, \mathbf{p}, \mathcal{R}) =\;
&\lambda_{\text{angle}}\, \mathcal{L}_{\text{angle}}(\theta) \\
&+ \lambda_{\text{overlap}}\, \mathcal{L}_{\text{overlap}}(\mathbf{s}, \mathbf{p}) \\
&+ \lambda_{\text{region}}\, \mathcal{L}_{\text{region}}(\mathbf{s}, \mathcal{R})
\end{align}
where $f_{\text{guidance}}(\mathbf{s}, \mathbf{p}, \mathcal{R}) \approx 0$ for valid configurations and increases as constraints are violated. Its gradient steers the diffusion denoising trajectory towards the feasible set, in the spirit of the energy-guided conditional diffusion of~\citep{yu2023freedom}. The $\mathbf{s}$ are the current states (e.g., positions and angles of pieces),
$\mathbf{p}$ are the piece shapes,
$\mathcal{R}$ is the target region,
$\lambda_{\text{angle}}, \lambda_{\text{overlap}}, \lambda_{\text{region}}$ are weighting coefficients.

\paragraph{Angle Loss (Soft-Min to Target Angles)}
\begin{equation}
    \mathcal{L}_{\text{angle}}(\theta) = \sum_{i=1}^N \left( -\frac{1}{\beta} \log \sum_{k} \exp\left[ -\beta (\theta_i - \theta_k^*)^2 \right] \right)
\end{equation}
where $\theta_k$ are the target angles (e.g., $0^\circ, 90^\circ, \ldots$), and $\beta$ controls the sharpness.

\paragraph{Rectangle Overlap Loss (Softplus Centre-Distance Penalty)}
\begin{equation}
    \mathcal{L}_{\text{overlap}}(\mathbf{s}, \mathbf{p}) = \frac{1}{2} \sum_{i \neq j} \text{softplus}\left( r_i + r_j - d_{ij} \right)
\end{equation}
where $d_{ij}$ is the distance between centres, and $r_i$ is the effective radius of rectangle $i$. For tangram, whose pieces are anisotropic polygons, we instead use the signed-distance polygon overlap penalty described above; the circular surrogate in this subsection is not used for tangram.

\paragraph{Region Loss (Softplus Boundary):}
\begin{align}
     &\mathcal{L}_{\text{region}}(\mathbf{s}, \mathcal{R})\notag\\= &\sum_{i=1}^N [ \text{softplus}(x_{\min} - x_i) + \text{softplus}(x_i - x_{\max}) \notag \\ & + \text{softplus}(y_{\min} - y_i) + \text{softplus}(y_i - y_{\max})]
\end{align}
where $(x_i, y_i)$ is the position of piece $i$, and $(x_{\min}, x_{\max}, y_{\min}, y_{\max})$ define the region.

\begin{table*}[htbp]
    \centering\small
    \caption{The detailed human study results (per-participant choice counts). The numbers are the choice counts for the generated image and the ground-truth image.}
    \label{tab:human_study_result_appendix}
    \begin{tabular}{@{}lccccc@{}}
    \toprule
    & Participant 1 & Participant 2 & Participant 3 & Participant 4 & Participant 5 \\ \midrule
    GAG MCTS (chosen)  & 19.0 & 21.0 & 16.0& 21.0 & 21.0 \\
    Auto-Regressive (chosen) & 3.0 & 1.0 & 4.0& 1.0 & 2.0 \\\bottomrule
    \end{tabular}
\end{table*}

\section{Evaluation metrics}\label{app:metrics}

We report three metrics on every Tangram Assembly experiment: FID, precision/recall, and validity.

\paragraph{FID.} The Fr\'{e}chet Inception Distance measures the distance between the Gaussian moments (mean and covariance) of feature representations of the real and generated samples, and is a standard likelihood proxy for generative image models. We compute FID with two feature extractors: a pre-trained InceptionV3 (``FID''), and a CLIP model fine-tuned on the KiloGram dataset (``FID$_\text{clip}$''). Because InceptionV3 is pre-trained on natural images, its features are not directly calibrated to the Tangram distribution; the CLIP variant complements it with a domain-adapted backbone.

\paragraph{Precision and recall.} Following~\citet{kynkaanniemi2019improved}, we measure diversity and fidelity via a precision/recall metric that quantifies the coverage of the real-data manifold by generated samples (precision) and vice versa (recall). The reported precision and recall values use the KiloGram-fine-tuned CLIP backbone with $k=3$ nearest neighbours. Images are processed to extract feature representations $f_r\in\mathbb{R}^d$ for real samples and $f_g\in\mathbb{R}^d$ for generated samples. For each real sample $i$, a neighbourhood radius $\tau_i^r$ is computed as the distance to its $k$-th nearest neighbour among the other real samples; for each generated sample $j$, $\tau_j^g$ is computed analogously among generated samples. Precision is the fraction of generated samples that fall within at least one real sample's neighbourhood, and recall is the fraction of real samples covered by at least one generated sample's neighbourhood:
\begin{align}
\text{Precision} &= \frac{1}{M} \sum_{j=1}^{M} \mathbb{I}\!\left[\exists\, i : \mathrm{dist}(f_g^j, f_r^i) \leq \tau_i^r\right], \\
\text{Recall} &= \frac{1}{N} \sum_{i=1}^{N} \mathbb{I}\!\left[\exists\, j : \mathrm{dist}(f_r^i, f_g^j) \leq \tau_j^g\right],
\end{align}
where $N$ and $M$ are the number of real and generated samples respectively and $\mathbb{I}[\cdot]$ is the indicator function. High precision with low recall indicates mode collapse; low precision with high recall indicates poor fidelity.

\paragraph{Validity.} We also report the percentage of generated configurations that satisfy the hard geometrical constraints (all seven pieces used, no pairwise overlap, connected contact graph), since FID and precision/recall operate in feature space and do not directly penalise constraint violations.

\section{Human Study Details}\label{app:human_study}

Our human study was conducted with 5 participants. We randomly sampled 50 pairs of generated and ground-truth tangram images with text descriptions, and let 5 participants choose which one matches the text description better. For each pair of images, we show both of them on the screen and ask the participants: ``Which image better matches the description xxx?''

\begin{figure}[htbp]
    \centering
    \includegraphics[width=\linewidth]{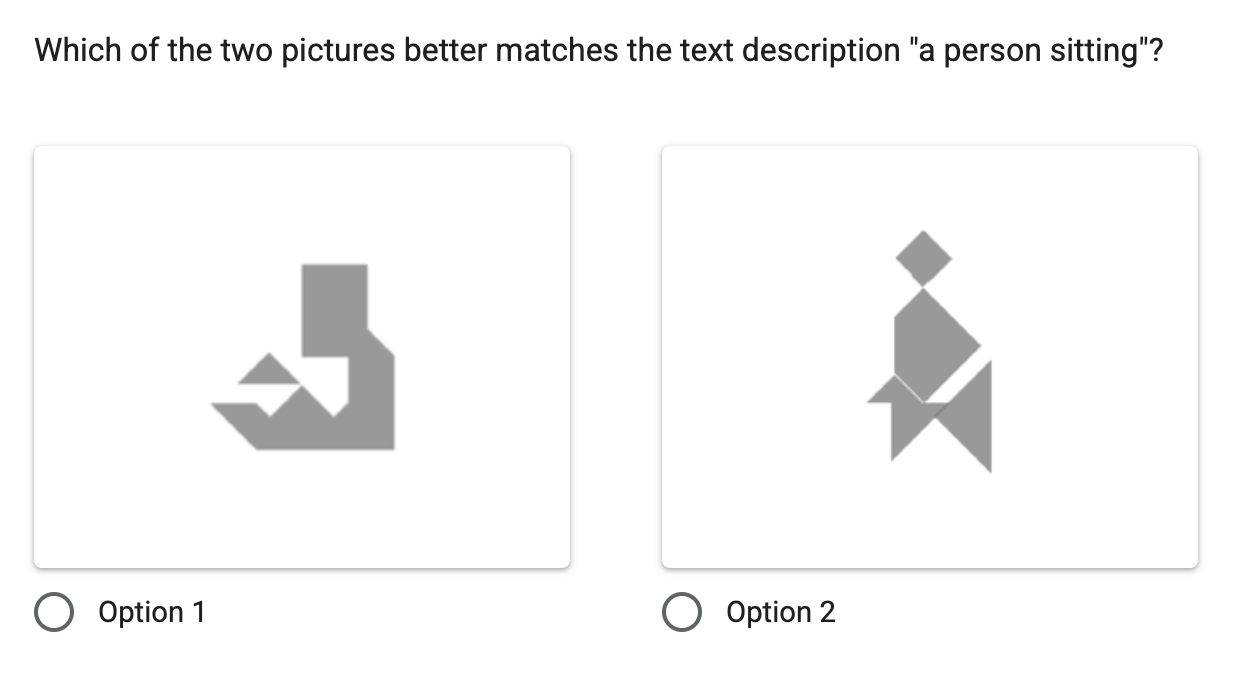} 
    \caption{One example question in the questionnaire. We randomly sampled 50 pairs of generated and ground-truth tangram images with text descriptions, and let 5 participants choose which one matches the text description better. For each pair of images, we show both of them on the screen and ask the participants: ``Which image better matches the description xxx?'' The placement of the image is random, i.e., the generated image can be placed on the left side or the right side.}
    \label{fig:human_study_example}
\end{figure}

One example question of the questionnaire is shown in Fig~\ref{fig:human_study_example}.

The detailed results are shown in Table~\ref{tab:human_study_result_appendix}, including the choice counts for the generated image and the ground-truth image. For each participant, we randomly shuffle the pairs of images and let them choose the better one. To test whether the difference in selection frequencies between GAG MCTS and the autoregressive baseline is statistically significant, we run a paired $t$-test on the per-participant preference rates relative to the common baseline (human-labelled images). The test yields $t(4)=11.33$, $p<0.001$, with Cohen's $d=5.07$, confirming a large and statistically significant effect.

\newpage
\section*{NeurIPS Paper Checklist}

\begin{enumerate}

\item {\bf Claims}
    \item[] Question: Do the main claims made in the abstract and introduction accurately reflect the paper's contributions and scope?
    \item[] Answer: \answerYes{}
    \item[] Justification: The abstract and introduction clearly state the scope (tangram assembly and rectangle composition as case studies for constrained generation) and the main claims (diffusion models struggle with hard constraints; autoregressive generation with RL and MCTS improves feasibility). Limitations are discussed in Section~\ref{sec:discussion}.

\item {\bf Limitations}
    \item[] Question: Does the paper discuss the limitations of the work performed by the authors?
    \item[] Answer: \answerYes{}
    \item[] Justification: Section~\ref{sec:discussion} discusses two main limitations: (1) the CLIP-based reward is weakly calibrated under limited KiloGram data, and (2) MCTS inference cost is roughly two orders of magnitude higher than single-pass baselines.

\item {\bf Theory assumptions and proofs}
    \item[] Question: For each theoretical result, does the paper provide the full set of assumptions and a complete (and correct) proof?
    \item[] Answer: \answerYes{}
    \item[] Justification: The feasible-mass bound (Proposition~1) is proved in Appendix~\ref{app:theory_statements}. The rectangle pose-volume decomposition is derived via the chain rule of conditional probability in Appendix~\ref{app:apply_rect}. The tangram tube-volume bound applies the per-stratum precision lemma of~\citet{soh2026hallucination} and the union bound (Appendix~\ref{app:apply_tangram}); all assumptions (uniform reference density, bounded layout space) are stated explicitly.

    \item {\bf Experimental result reproducibility}
    \item[] Question: Does the paper fully disclose all the information needed to reproduce the main experimental results of the paper to the extent that it affects the main claims and/or conclusions of the paper (regardless of whether the code and data are provided or not)?
    \item[] Answer: \answerYes{}
    \item[] Justification: All training details (network architectures, hyperparameters, optimisers, MCTS simulation budgets, adversarial training loop) are described in Section~\ref{sec:method} and Appendix~\ref{app:ch5_implementation_details}. The dataset (KiloGram) is publicly available. The rectangle packing dataset generation procedure is described in Appendix~\ref{app:ch5_rectangle_gen}.

\item {\bf Open access to data and code}
    \item[] Question: Does the paper provide open access to the data and code, with sufficient instructions to faithfully reproduce the main experimental results, as described in supplemental material?
    \item[] Answer: \answerNo{}
    \item[] Justification: The KiloGram dataset is publicly available at \url{https://lil.nlp.cornell.edu/kilogram/}, and the rectangle packing dataset can be regenerated with the procedure described in Appendix~\ref{app:ch5_rectangle_gen}. The codebase is adapted from the open-source Mctx library (\url{https://github.com/google-deepmind/mctx}) but is not yet released; we plan to release it upon acceptance.

\item {\bf Experimental setting/details}
    \item[] Question: Does the paper specify all the training and test details (e.g., data splits, hyperparameters, how they were chosen, type of optimizer) necessary to understand the results?
    \item[] Answer: \answerYes{}
    \item[] Justification: Section~\ref{sec:experiments} describes the experimental setup (data splits, evaluation metrics, baselines). Full hyperparameters, optimiser settings, and architecture details are provided in Appendix~\ref{app:ch5_implementation_details}.

\item {\bf Experiment statistical significance}
    \item[] Question: Does the paper report error bars suitably and correctly defined or other appropriate information about the statistical significance of the experiments?
    \item[] Answer: \answerNo{}
    \item[] Justification: Error bars are not reported for the main results because each training run (MCTS self-play with adversarial reward refinement) is computationally expensive. The human study reports mean $\pm$ standard deviation across participants. We evaluate on the full test set to reduce variance from the evaluation side.

\item {\bf Experiments compute resources}
    \item[] Question: For each experiment, does the paper provide sufficient information on the computer resources (type of compute workers, memory, time of execution) needed to reproduce the experiments?
    \item[] Answer: \answerYes{}
    \item[] Justification: Appendix~\ref{app:inference_cost} reports wall-clock inference times per configuration and specifies the hardware used: Intel Xeon Platinum 8580 CPU with 192 CPU cores available and 2 NVIDIA A100 PCIe GPUs with 80GB memory per GPU. Appendix~\ref{app:ch5_implementation_details} reports batch sizes and optimisation details.
    
\item {\bf Code of ethics}
    \item[] Question: Does the research conducted in the paper conform, in every respect, with the NeurIPS Code of Ethics \url{https://neurips.cc/public/EthicsGuidelines}?
    \item[] Answer: \answerYes{}
    \item[] Justification: The research uses publicly available datasets and open-source code. The human study is a small-scale qualitative sanity check with no risk to participants.

\item {\bf Broader impacts}
    \item[] Question: Does the paper discuss both potential positive societal impacts and negative societal impacts of the work performed?
    \item[] Answer: \answerNA{}
    \item[] Justification: This is foundational research on constrained generation methods studied through abstract puzzle domains (tangram and rectangle composition). We do not foresee direct negative societal impacts from this work.
    
\item {\bf Safeguards}
    \item[] Question: Does the paper describe safeguards that have been put in place for responsible release of data or models that have a high risk for misuse (e.g., pre-trained language models, image generators, or scraped datasets)?
    \item[] Answer: \answerNA{}
    \item[] Justification: The models are task-specific (tangram and rectangle layout generation) and pose no risk for misuse.

\item {\bf Licenses for existing assets}
    \item[] Question: Are the creators or original owners of assets (e.g., code, data, models), used in the paper, properly credited and are the license and terms of use explicitly mentioned and properly respected?
    \item[] Answer: \answerYes{}
    \item[] Justification: The KiloGram dataset~\citep{ji2022abstract} is public on Hugging Face (\url{https://huggingface.co/datasets/lil-lab/kilogram}) and its GitHub repository states an MIT license for the dataset and code, with Tangram images available for educational and research use under the linked Princeton Library notice. The Mctx library (\url{https://github.com/google-deepmind/mctx}) is licensed under Apache License 2.0. The pre-trained CLIP, ViT, and BERT assets are credited and use public model or repository pages with MIT (CLIP) or Apache-2.0 (ViT and BERT) licensing.

\item {\bf New assets}
    \item[] Question: Are new assets introduced in the paper well documented and is the documentation provided alongside the assets?
    \item[] Answer: \answerNA{}
    \item[] Justification: The paper does not release new datasets or pre-trained models. The rectangle packing dataset is procedurally generated and the generation procedure is described in Appendix~\ref{app:ch5_rectangle_gen}.

\item {\bf Crowdsourcing and research with human subjects}
    \item[] Question: For crowdsourcing experiments and research with human subjects, does the paper include the full text of instructions given to participants and screenshots, if applicable, as well as details about compensation (if any)?
    \item[] Answer: \answerYes{}
    \item[] Justification: The human study protocol, instructions, and per-participant results are described in Section~\ref{sec:exp_tangram} and Appendix~\ref{app:human_study}. The study is a small-scale qualitative sanity check (5 participants, 50 pairs each) used as a supportive signal, not a main result.

\item {\bf Institutional review board (IRB) approvals or equivalent for research with human subjects}
    \item[] Question: Does the paper describe potential risks incurred by study participants, whether such risks were disclosed to the subjects, and whether Institutional Review Board (IRB) approvals (or an equivalent approval/review based on the requirements of your country or institution) were obtained?
    \item[] Answer: \answerNA{}
    \item[] Justification: The human study is a small-scale qualitative sanity check (5 participants comparing abstract tangram images) with no risk to participants. No personal data is collected and the task poses no physical or psychological risk, so IRB approval is not required.

\item {\bf Declaration of LLM usage}
    \item[] Question: Does the paper describe the usage of LLMs if it is an important, original, or non-standard component of the core methods in this research? Note that if the LLM is used only for writing, editing, or formatting purposes and does \emph{not} impact the core methodology, scientific rigor, or originality of the research, declaration is not required.
    \item[] Answer: \answerNA{}
    \item[] Justification: LLMs are not used as a component of the core methods in this research.

\end{enumerate}

\end{document}